\pdfoutput=1

\documentclass[11pt]{article}

\usepackage[final]{acl}

\usepackage{times}
\usepackage{latexsym}
\usepackage{xurl}

\usepackage[T1]{fontenc}

\usepackage[utf8]{inputenc}

\usepackage{microtype}

\usepackage{inconsolata}

\usepackage{graphicx}

\usepackage{enumitem,amssymb}
\usepackage{amsmath}
\usepackage{wrapfig}
\usepackage{subcaption}
\usepackage{enumitem}
\usepackage{array}
\usepackage{multirow}
\usepackage{booktabs}
\usepackage{algorithm}
\usepackage{algpseudocode}
\usepackage[color=lightgray!20,textsize=scriptsize]{todonotes}

\title{Analyzing (In)Abilities of SAEs via Formal Languages}

\author{Abhinav Menon* \\ IIIT Hyderabad \\ \texttt{abhinav.m@research.iiit.ac.in}
        \And Manish Shrivastava \\ IIIT Hyderabad \\ \texttt{m.shrivastava@iiit.ac.in}
        \AND
        David Krueger \\ MILA \\ \texttt{david.scott.krueger@gmail.com} \And Ekdeep Singh Lubana* \\ CBS, Harvard University \\ \texttt{ekdeeplubana@fas.harvard.edu}
}

\begin{document}
\maketitle
\begin{abstract}
\vspace{-5pt}
Autoencoders\def\thefootnote{*}\footnotetext{These authors contributed equally.} have been used for finding interpretable and disentangled features underlying neural network representations in both image and text domains.
While the efficacy and pitfalls of such methods are well-studied in vision, there is a lack of corresponding results, both qualitative and quantitative, for the text domain.
We aim to address this gap by training sparse autoencoders (SAEs) on a synthetic testbed of formal languages.
Specifically, we train SAEs on the hidden representations of models trained on formal languages (Dyck-2, Expr, and English PCFG) under a wide variety of hyperparameter settings, finding interpretable latents often emerge in the features learned by our SAEs.
However, similar to vision, we find performance turns out to be highly sensitive to inductive biases of the training pipeline.
Moreover, we show latents correlating to certain features of the input do not always induce a causal impact on model's computation.
We thus argue that causality has to become a central target in SAE training: learning of causal features should be incentivized from the ground-up. 
Motivated by this, we propose and perform preliminary investigations for an approach that promotes learning of causally relevant features in our formal language setting.
\end{abstract}

\section{Introduction}

\begin{figure}
\centering
\vspace{-18pt}
\includegraphics[width=0.9\linewidth]{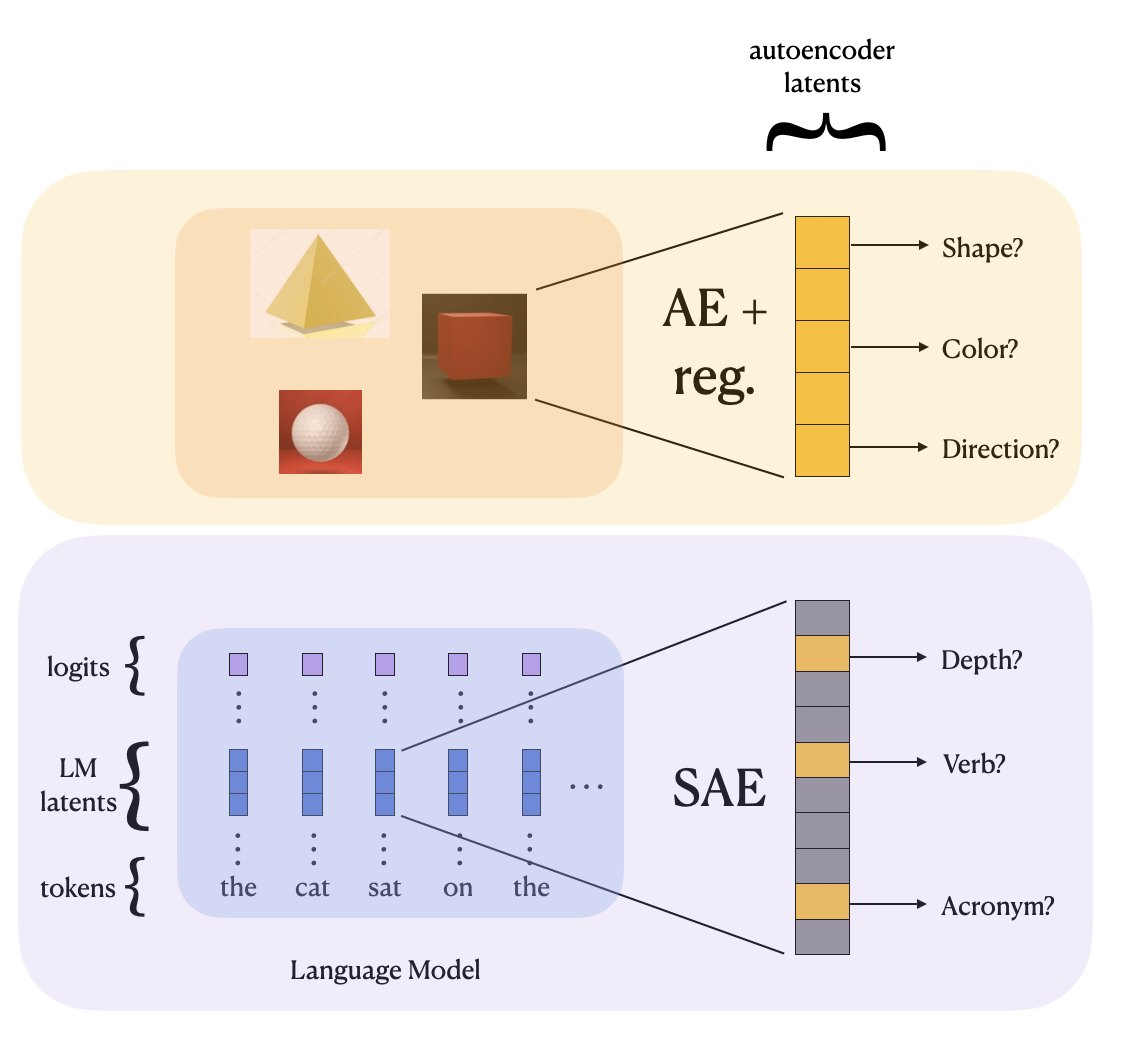}
\vspace{-5pt}
\caption{\textbf{The autoencoder paradigm for interpretability.}
Autoencoders have formed the basis of approaches to disentanglement in the vision domain (\citealp{higgins2016beta}). Utilizing synthetic testbeds, prior work has shown several limitations in this general pipeline~\cite{locatello2019challenging, trauble2021disentangled}. While SAEs have similarly been used to disentangle hidden representations of language models for interpretability, we aim to perform a similar study as ones in vision to understand the (in)abilities of SAEs.
\vspace{-12pt}
}
\label{fig:intro}
\end{figure}

In recent years, mechanistic interpretability has gained currency as an approach towards understanding the functioning of language models (LMs)~\cite{nanda2023progress, olsson2022context, wang2022interpretability, li2024inference}. 
A mechanistic interpretability paradigm that has seen remarkable progress especially is sparse autoencoders (SAEs), which aims to disentangle model representations into independent, interpretable components~\cite{lieberum2024gemma, bricken2023towards, cunningham2023sparse, engels2024not, gao2024scaling, marks2024sparse, rajamanoharan2024improving} (see App.~\ref{sec:related_app} for related work).
Similar interpretability approaches based on autoencoders have been extensively studied in the vision domain (see Fig.~\ref{fig:intro}), from both empirical and theoretical viewpoints~\cite{locatello2019challenging, trauble2021disentangled, lachapelle2023synergies, locatello2020weakly, klindt2020towards}. 
Specifically, using \textit{synthetic testbeds} where a ground-truth set of latents underlying the data-generating process can be listed~\cite{higgins2016beta}, these studies have established results on the (non-)identifiability of disentangled representations via autoencoders and argued for the importance of inductive biases, which motivate different regularization terms.
For instance, \citet{locatello2019challenging} prove that for any given dataset (modeled as a multivariate distribution), there are infinitely many generative models, all entangled with each other, and remark that inductive biases are necessary to ``select'' from among these distributions.
\citet{trauble2021disentangled} show that in a correlated dataset (which real-world datasets tend to be), a perfectly disentangled model would lead to suboptimal log-likelihood, and therefore would not be learned without significant inductive biases.
%
We argue a similar study is needed for understanding the (in)abilities of SAEs for interpreting LM representations.

\textbf{This work.} We propose the use of synthetic \textit{language} testbeds to stress-test the SAE approach to interpretability of LMs.
In particular, we train a spectrum of SAEs on a set of formal languages---specifically, probabilistic context free grammars (Dyck-2, Expr, and English)---and demonstrate interpretable latents that activate for relevant concepts of the data-generating process (e.g., parts-of-speech) are easily identifiable.
However, similar to vision, we find results are highly sensitive to the training pipeline, with identified latents rarely having a causal impact on the model output.
This indicates new approaches that directly target learning of causally relevant latents are needed. 
To take a step in this direction, we thus propose a training pipeline that exploits token-level correlations as ``weak supervision''~\citep{locatello2020weakly}.
Overall, our contributions can be summarized as follows.

\begin{itemize}[leftmargin=14pt, itemsep=1pt, topsep=2pt, parsep=1pt, partopsep=1pt]
    \item We define formal languages of different complexities, train Transformers on these languages, and then analyze SAEs trained on their representations under several different hyperparameter settings. We \textbf{identify several features} occurring in these SAEs that correspond to variables central to the data generating process; e.g., depth and part of speech (Sec.~\ref{sec:experiments}). 
    
    \item We demonstrate, in line with results from vision, that identifiability of disentangled features is \textit{not} robust (Sec.~\ref{sec:results}). Results are \textbf{highly sensitive} to changes in normalization methods, hyperparameters, and other such settings. For example, models trained under $L_1$ regularization consistently fail to find interpretable features; additionally, normalization has widely varying effects on the reconstruction abilities. 
    
    \item We demonstrate that mere identification of \textbf{disentangled features does not imply said features are causally relevant} to the model's computation (Sec.~\ref{sec:lack_causality}). This indicates causality should be actively modeled as a constraint in the training pipeline of SAEs. We take a step towards this objective by proposing a modified SAE training protocol that deems correlations between tokens as naturally available interventions, similar to use of correlated images in autoencoder based approach for vision interpretability~\citep{locatello2020weakly, klindt2020towards, brehmer2022weakly}. Results show that the proposed protocol often yields latents with interpretable and causal impact on model outputs in our formal language setting (Sec.~\ref{sec:causality}). These results can be deemed a proof-of-concept demonstration that insights from prior work on autonencoder-based interpretability may continue to be useful for LM settings as well.
\end{itemize}

\section{Experimental Setup}
\label{sec:experiments}
Our experiments consist of training SAEs (of various paradigms) on the intermediate representation of transformer models trained on formal languages. 
We explain the data generating process, model architectures, and training paradigms next.

\subsection{Data and Models}
The formal languages we work with are probabilistic context-free grammars (PCFGs), which are generated by starting with a fixed `start' symbol, and probabilistically replacing nonterminals according to production rules. 
We work with three PCFGs, intended to represent levels of complexity (in parsing and generation). 
In order of increasing complexity, the languages we consider are \textbf{Dyck-2} (the language of all matched bracket sequences with two types of brackets), \textbf{Expr} (a language of prefix arithmetic expressions), and \textbf{English} (a simple fragment of English syntax with only subject-verb-object constructions).
See App.~\ref{sec:grammars_app} for precise details.

We train Transformers~\cite{karpathy2022nanogpt} via the standard autoregressive language modeling pipeline on each of the above languages. 
%
%
The models have 128-dimensional embeddings, with 4 attention heads and MLPs, and 2 blocks. 
In all cases, the models achieve more than 99\% validity, i.e., under stochastic decoding, the strings generated belong to the language more than 99\% of the time.

\subsection{SAE Architecture}
\label{sec:sae}
Broadly, we use the conventional SAE architecture, which consists of two linear layers (an encoder and a decoder transform) with an activation function in between. More explicitly, if the sizes of the input and hidden state are $d$ and $h$ respectively, then our SAEs implement functions of the type
$$\operatorname{SAE}(x) = W_\text{dec}(\sigma(W_\text{enc}(x) + b_\text{enc})) + b_\text{dec},$$
where $W_\text{enc} \in \mathbb{R}^{h \times d}$, $W_\text{dec} \in \mathbb{R}^{d \times h}$, and $b_\text{enc} \in \mathbb{R}^h, b_\text{dec} \in \mathbb{R}^d$. We pick $h$ from $\{d, 2d, 4d, 8d\}$ and $\sigma = \text{ReLU}$ as the activation function. 
We refer to the encoder transform (up to and including the nonlinearity) as $E$, the decoder as $D$, and the hidden representation (the output of the encoder) as a \textbf{latent}. When these latents have interpretable explanations, identified correlationally (Sec.~\ref{sec:features}), we also refer to them as \textbf{features}.

Training is performed by optimizing on the MSE between $x$ and $\operatorname{SAE}(x)$, where the input $x$ to the SAEs is the output of the first block of the Transformer model.
Sparsity is enforced via two main methods. The first (and most common in existing work) is an $L_1$-regularization term added to the loss, which encourages latent representations to have low $L_1$-norm. The hyperparameter for this method is the weight of the regularization term. 
We also use, following \citet{gao2024scaling}, top-$k$ regularization---at the SAE hidden layer, after the activation, we select the highest $k$ latents, and zero out the rest. Effectively, we force the $L_0$-norm of the latents to be at most $k$, i.e., the hyperparameter is $k$ itself. We examine the effects of hyperparameters on the performance of SAEs in Sec.~\ref{sec:sensitivity}.
We analyze variants of the architecture above by altering the normalization method and pre-bias. 
There are three parts of the operation of the SAE that can be normalized---the input, the reconstruction, and the decoder directions. 
This yields four architecture variants: \textit{no normalization \textbf{(I)}, input and decoder with pre-bias \textbf{(II)}, input alone \textbf{(III)}, and input and reconstruction \textbf{(IV)}.}
By pre-bias, we refer to the addition of a learnable vector $b_\text{pre} \in \mathbb{R}^d$ to the input before applying the encoder, which is then subtracted after applying the decoder.

\subsection{Causality of Features}
\label{sec:exp_causality}
Given a trained SAE$: x \mapsto D(E(x))$, we examine the causality of the latents by defining a \textit{reconstructed run} of the transformer. Let the model's two layers be $L_0$ and $L_1$; then a normal run is\vspace{-2pt}
$\text{logits} = W_\text{LM}L_1(L_0(t)),$
where $t \in \mathbb{R}^{s \times d}$ represents the embeddings of a sequence of $s$ tokens, and $W_\text{LM} \in \mathbb{R}^{d \times |V|}$ is the projection of the final representations into the logit space. The prediction of the next token is given, therefore, by $\operatorname{softmax}(\text{logits})$. 
We then define a reconstructed run as
$\text{logits} = W_\text{LM}L_1(D(E(L_0(t))));$
i.e., we interrupt the run after the first layer, pass the output through the SAE, and resume the run using the reconstruction of the SAE.
For ease of notation, we partition the reconstructed run into two functions, which chained together give the complete run:
\begin{align*}
    f := E \circ L_0; \quad g := W_\text{LM} \circ L_1 \circ D.
\end{align*}
Thus, the latent of a token $x$ is given by $f(x)$ and the logit distribution of a token with latent $l$ is given by $g(l)$.
To study the causality of the representations, then, we intervene on the \textit{latent representation}, between $E$ and $D$ (in other words, between $f$ and $g$). We intervene on the specific elements of the latent that correlate with interpretable features of the input tokens; for instance, in the English grammars, we search for latents correlating with each part of speech in the grammar. 
Our interventions consist of `clamping' latents to some fixed value. In other words, for each token, we run the forward pass and extract the latent representation; we set certain elements to a clamp value, leaving the rest unchanged---this value is then passed through the decoder $D$ and the rest of the forward pass.

We examine the effects of the interventions at the sentence level (Sec.~\ref{sec:lack_causality}). This is done in the context of free generation, where the model is prompted by a \texttt{<BOS>} token, and the above intervention is run on \textit{each token's} representation. We intervene using values spaced at 10 equal intervals in $[-v_\text{max}, v_\text{max}]$, where $v_\text{max}$ is the maximum absolute value attained by the intervened feature. We use both positive and negative values to account for the possibility that the feature may be antipodal to our explanation.

In Sec.~\ref{sec:causality}, we incentivize the SAEs to identify features with causal function by adding a regularization term (Fig.~\ref{fig:causal_loss}). 
We carry out the same studies noted above for those SAEs and qualitatively connect them to the nature of regularization (Sec.~\ref{sec:causal_analysis}).

\section{Results}
\label{sec:results}

\begin{figure}
  \centering
  \includegraphics[width=0.9\linewidth]{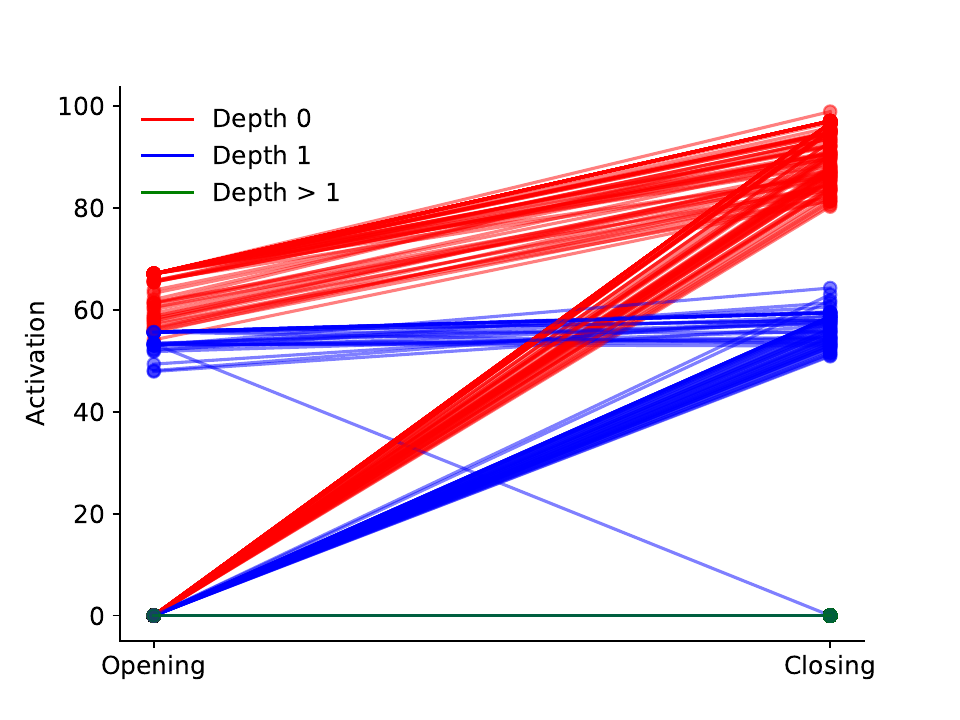}
  \caption{\textbf{A feature matching corresponding opening and closing brackets.} Each line represents a pair of brackets, and joins the opening bracket's activation (left) to the closing bracket's (right). We note that the depth and opening activation are sufficient to determine the closing activation, and that the opening and closing activations are sufficient to determine the depth.}
  \label{fig:matching}
\end{figure}

Qualitatively, we observe that top-$k$-regularized SAEs tend to have more interpretable features than $L_1$-regularized ones. 
The former results in interpretable features across all languages---we find features representing fundamental aspects of the corresponding grammars, as discussed next briefly. 
See App.~\ref{sec:features_app} for a description of our feature identification pipeline, and further results, including other features we are able to identify and how strongly features correlate with their claimed explanations.

In the case of Dyck-2, we expect the depth (the number of brackets yet to be closed) to be represented. We find a feature that \textit{thresholds} the depth of tokens, i.e., it activates on tokens with depth above a certain threshold depth $D$. Usually, $D$ is greater than the mean; for instance, when mean depth is 8.3, we find $D = 11$. We also find features that match corresponding pairs of opening and closing brackets, usually at lower depths (Fig.~\ref{fig:matching}). These features take values from only a few small ranges, and their values at an opening bracket determine those they take at the matching closing bracket. 
These features, even though they are binary, indicate that the token representations do maintain some form of counter (which aligns with claims from prior work, like~\citet{bhattamishra2020ability}). 

\begin{figure}
\centering
  \includegraphics[width=0.85\linewidth]{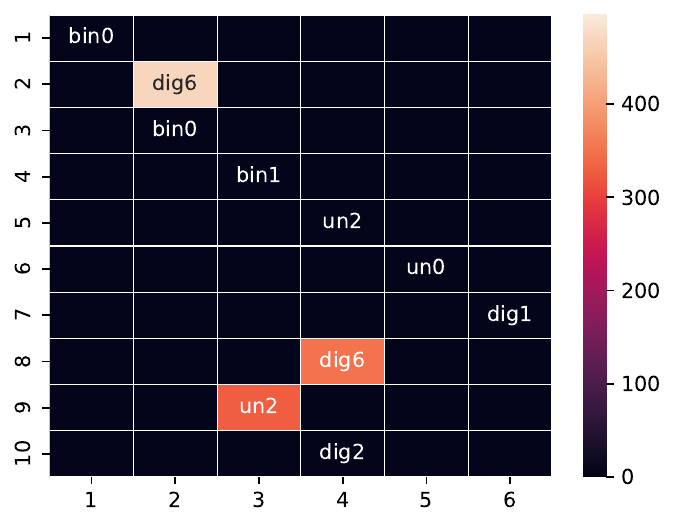}
  \caption{\textbf{A feature that activates when exactly one more expression is required.} Here, the x-axis is token depth, and the y-axis is token index. The lines connect the operators to their operands.}
  \label{fig:almost}
\end{figure}

\begin{figure*}
    \includegraphics[width=0.95\textwidth]{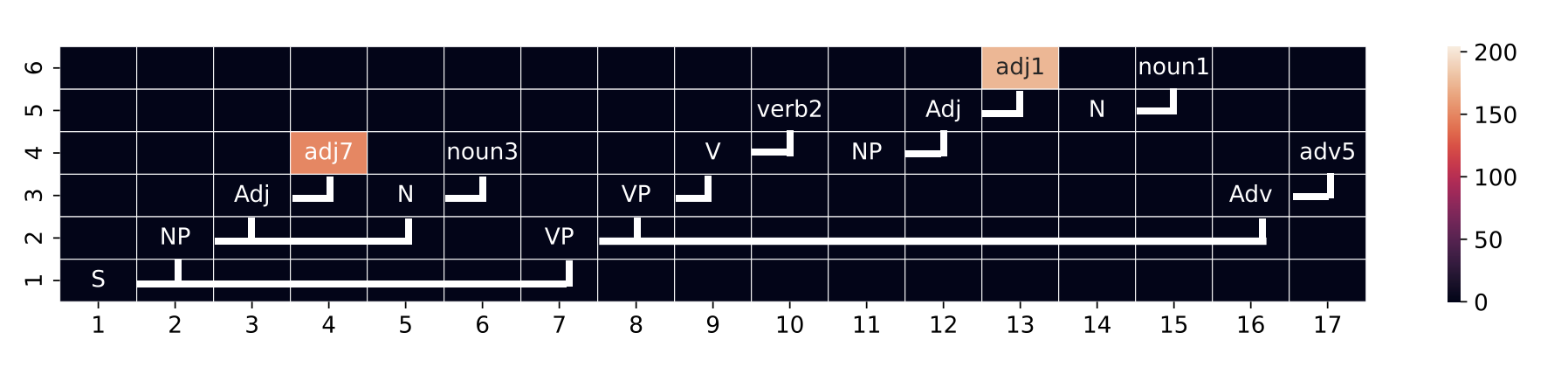}
    \vspace{-10pt}
    \caption{\textbf{A feature that activates only on adjectives, at any position.} Here, depth is represented by the y-axis and position by the x-axis; the lines connect nonterminals to their productions (see App.~\ref{sec:grammars_app} for the production rules). The cell color represents the activation magnitude.}
    \label{fig:adjectives}
\end{figure*}

In Expr, the analogue of depth is a counter that indicates how many more expressions are needed to complete the sequence (App.~\ref{sec:grammars_app}). We find a feature that activates exactly when this counter's value is 1, i.e., when exactly one expression is required (Fig.~\ref{fig:almost}). This provides strong evidence of a counter process being implemented; in fact, we note that generation without this process would be rather convoluted (Algorithm~\ref{alg:almost_done}).

An inference we can make from the Expr features is that there is an implicit ``type'' feature in the representations, distinguishing operators of different valences. A natural place to look for this is the parts of speech in our English grammar, and we find that $k$-regularized SAEs do contain features corresponding to each part of speech. For example, we illustrate the `adjective' feature in Fig.~\ref{fig:adjectives}.

Further, we note a scaling relation in the performance of the top-$k$-regularized SAEs with hidden dimension size. The reconstruction loss they achieve (after a fixed number of iterations) decreases according to a power law with the size of the autoencoder's hidden layer (see App.~\ref{sec:power_law}); similar results were seen by \citet{sharkey2024takingout}, relating the reconstruction loss to the L1 penalty coefficient, indicating our setup captures part of the phenomenology observed with complex settings.

\begin{table*}
\footnotesize
\centering
    \begin{tabular}{lcccccccc}
\hline
\multirow{2}{*}{\textbf{Language}} & \multicolumn{4}{c}{$L_1$} & \multicolumn{4}{c}{top-$k$} \\
        \cmidrule(lr){2-5} \cmidrule(lr){6-9}
        & \textbf{I} & \textbf{II} & \textbf{III} & \textbf{IV} & \textbf{I} & \textbf{II} & \textbf{III} & \textbf{IV} \\
        \hline
        Dyck-2  &  0.01 & 0.0  & 0.01 & 0.0  & 49.27 &  3.48 & 50.02 & 0.06 \\
        Expr    & 33.31 & 6.50 & 0.06 & 0.49 & 99.88 & 69.14 & 99.76 & 0.0  \\
        English &  0.29 & 1.13 & 0.01 & 0.01 & 92.46 & 50.12 & 80.79 & 0.37 \\
        \bottomrule
    \end{tabular}
  \caption{\textbf{Sensitivity to Hyperparameters.} Reconstruction Accuracy (\%) averaged over regularization values and hidden size. We present the accuracies for SAEs with no normalization (I); with inputs and decoder normalized, and pre-bias (II); with inputs normalized (III); and with inputs and reconstructions normalized (IV). The reconstruction capabilities of the models shows high variance across hyperparameter settings.}
  \label{table:k-vs-alpha}
\end{table*}

\begin{table*}[!ht]
\footnotesize
\centering
\begin{tabular}{>{\centering\arraybackslash}p{8.2cm}|>{\centering\arraybackslash}p{1.6cm}|>{\centering\arraybackslash}p{1cm}|>{\centering\arraybackslash}p{1cm}|>{\centering\arraybackslash}p{1cm}}

\toprule
\textbf{Input Sequence} & \textbf{\#Required} & \textbf{Clamp} $-v_\text{max}$ & \textbf{Clamp} 0 & \textbf{Clamp} $v_\text{max}$ \\
\midrule
\texttt{un2 tern1 dig8 bin0 tern2 bin2 bin2 dig6 dig7 dig5} & 4 & 4 & 4 & 4 \\
\texttt{bin1 un1 tern1 bin2 dig0 dig7 dig0 bin1 dig3 dig7} & 1 & 1 & 1 & 1 \\
\texttt{un0 tern1 dig9 dig7 un0 tern1 un2 bin2 dig6 dig6} & 2 & 2 & 2 & 2 \\
\texttt{tern1 dig1 dig7 tern1 tern1 dig7 tern1 tern1 un0 un1} & 8 & 8 & 8 & 8 \\
\bottomrule
\end{tabular}
\vspace{-5pt}
\caption{\textbf{Behavior of the Expr model under interventions.} The `clamp' columns indicate the number of expressions generated by the model after being prompted by the input sequence and an intervention defined by the clamp value. We see that there is no effect of the intervention on the behavior of the model.}
\label{table:expr_intervention}
\end{table*}

\subsection{Are SAEs Robust to Hyperparameters?}
\label{sec:sensitivity}
As noted before, we find top-$k$-regularized SAEs consistently yield more interpretable features languages than $L_1$-regularized SAEs. We also measure the \textit{reconstruction accuracy}, or the percentage of valid generations of the model after the latents are substituted with the SAE reconstructions.
Tab.~\ref{table:k-vs-alpha} shows the average results for each combination of regularization, pre-bias, and normalization strategy (averaged across regularization values and hidden sizes). 
We see that top-$k$-regularized SAEs usually (but not always) outperform $L_1$-regularized ones if all other settings are kept the same. 
We also note from this table the high sensitivity to hyperparameter settings that SAEs exhibit: no clear trend is visible across languages, regularization methods, or normalization settings---in line with \citet{locatello2019challenging}'s findings.

\subsection{Does Correlation Imply Causation?}
\label{sec:lack_causality}
An interesting aspect of the features we identify is that despite strongly correlating with the discussed explanation (see Tab.~\ref{table:features}), they do \textit{not} have the expected causal effects. In this section, we outline our findings in causal experiments (Sec.~\ref{sec:exp_causality}) on the features described above.

\begin{figure}
    \vspace{-8pt}
    \centering
    \begin{subfigure}{0.5\linewidth}
    \includegraphics[width=\linewidth]{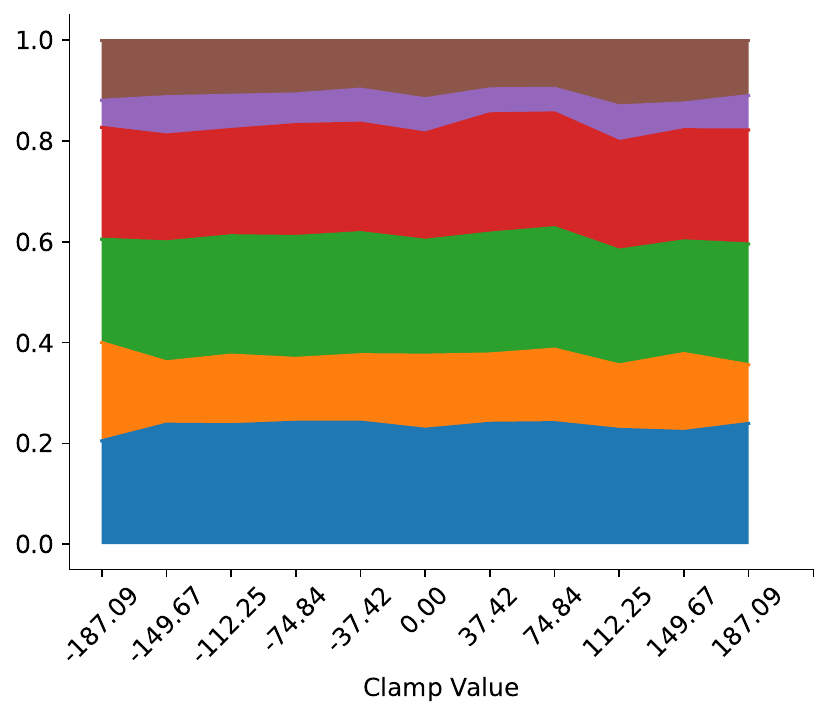}
    \end{subfigure}
    \begin{subfigure}{0.48\linewidth}
        \includegraphics[width=\linewidth]{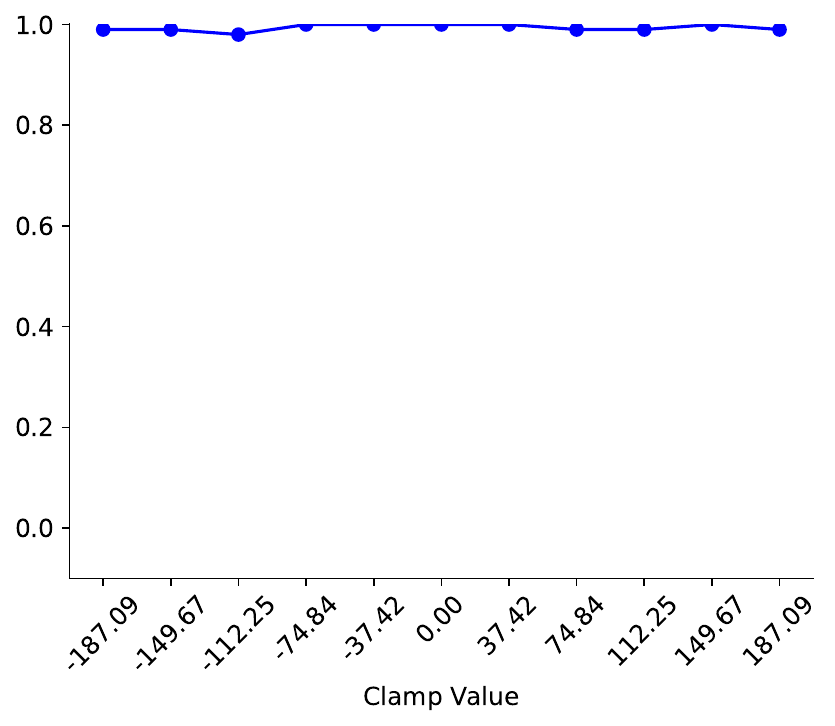}
    \end{subfigure}
    \caption{\textbf{Behavior of the English model under interventions.} We intervene on the model by replacing its hidden representations with the SAE's reconstructions, where an SAE latent (specifically, one corresponding to adjectives) is clamped to a fixed value. These values are selected at uniform intervals from $[-v_\text{max}, v_\text{max}]$, where $v_\text{max}$ is the maximum value taken by that latent (in line with \citet{templeton2024scaling}). For each value (x-axis), we plot the fraction of each part of speech (\textcolor{blue}{nouns}, \textcolor{orange}{pronouns}, \textcolor{green}{adjectives}, \textcolor{red}{verbs}, \textcolor{violet}{adverbs}, and \textcolor{brown}{conjunctions}) in the output (\textbf{left}) and the fraction of outputs that are grammatical (\textbf{right}). We see interventions yield essentially no visible effects.}
    \label{fig:intervention}
\end{figure}

In the case of Expr, we have seen in Sec.~\ref{sec:features} that there is a counter feature that activates when exactly one expression is left to complete the sequence. 
We pass an incomplete sequence to the model, and intervene on this feature by clamping it to one of $\{-v_\text{max}, 0, v_\text{max}\}$, where $v_\text{max}$ is the maximum absolute value of the latent (see Sec.~\ref{sec:exp_causality}). 
We then examine the generations that result in each case. 
We expect that high clamp values will cause the model to generate only one more expression (even if more may be required) and low values will cause it to generate more than one, or perhaps end the sequence immediately (even if exactly one is needed); or vice versa, since we cannot be certain of the direction to apply intervention in. 
Tab.~\ref{table:expr_intervention} presents the results of this experiment. 
For each input sample, we mention the number of expressions required to complete it, and the number of expressions actually generated by the model under each intervention.
\textit{We do not see the expected---or indeed any---causal effect of intervening on this feature---it appears that this feature is only correlational in nature.}

Similarly, we examine our English grammar. Here, we examine the effects of intervening on the part-of-speech features described in Sec.~\ref{sec:features}. Note that for a feature correlating with, say, adjectives, we can expect that if it has a causal effect, it must control the \textit{next-token} distribution for that part-of-speech (as this is the only task the model is trained on).
We thus apply our causality protocol (Sec.~\ref{sec:exp_causality}) to the feature correlating with adjectives. The results of this experiment are shown in Fig.~\ref{fig:intervention}.
\textit{In this case, as in Expr, we do not see a causal effect of intervening on this feature.} See App.~\ref{sec:causality_app} for similar results on other parts-of-speech.

\section{Incentivizing Causality}
\label{sec:causality}
\vspace{-5pt}

\begin{figure*}[!tpbh]
    \centering
    \vspace{-15pt}
    \includegraphics[width=0.88\linewidth]{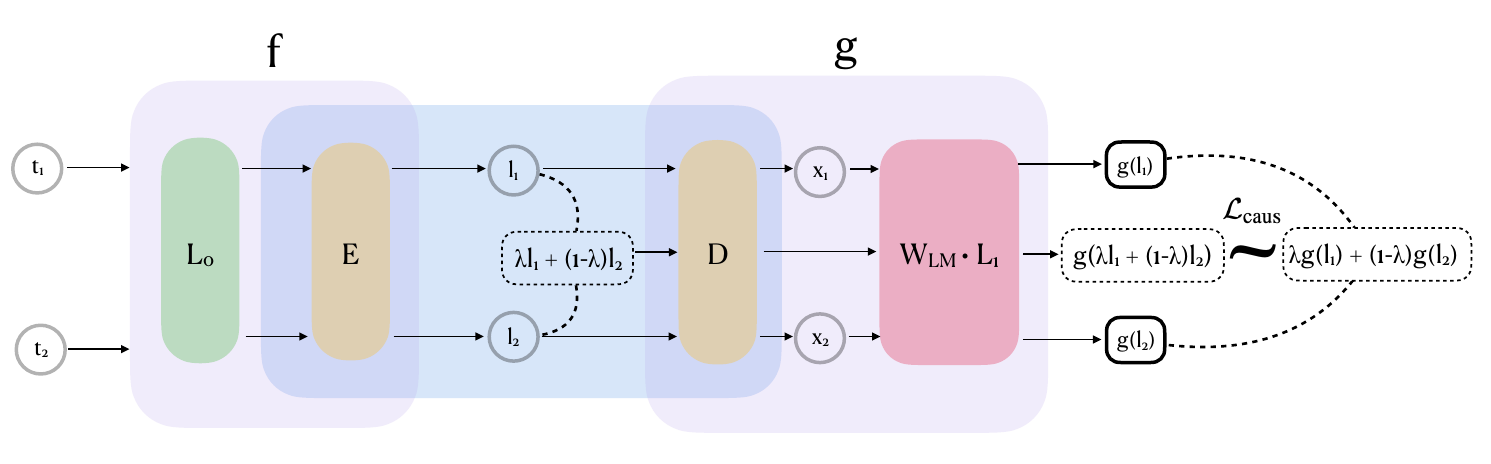}
    \caption{\textbf{Operationalizing the causal regularization term.} A clean run of the model consists of applying $L_0$, followed by $L_1$ and $W_\text{LM}$. We define a \textit{reconstructed run}, which introduces an SAE between these two layers. The SAE consists of an encoder $E$ and a decoder $D$. We denote the input embeddings as $t$, SAE latents as $l$, and reconstructed model activations as $x$. Furthermore, we group the first part of the reconstructed run as $f := E \circ L_0$, and the second part as $g := W_\text{LM} \circ L_1 \circ D$. Given two tokens $t_1$ and $t_2$, we interpolate between the latents $l_1$ and $l_2$ (indicated by a dotted line) and pass this as input to $g$; our causal loss $\mathcal{L}_\text{caus}$ is then given by the difference between the interpolation of the outputs, and the output of the interpolation.
    }
    \label{fig:causal_loss}
\end{figure*}

Our results above indicate that though SAEs are competent at identifying features correlated with semantically meaningful concepts, these features may not have the expected (or any) causal effects. 
This mirrors similar observations made in vision, where it has been noted that mere data reconstruction pipelines (e.g., based on autoencoders) can fail to disentangle the latent factors of a generative model~\citep{locatello2019challenging}. 
However, later work on the topic demonstrated that correlations in the data-generating process (e.g., correlation in nearby frames of a video) can be leveraged to obtain disentangled representations~\citep{locatello2020weakly, klindt2020towards}. 
As this method relies on supervision via other data samples, rather than a ground truth, it is described as \textbf{weak supervision}.
%
A warranted question then is whether weak supervision can be elicited in language modeling scenarios too.

Motivated by the above, we take inspiration from \citet{klindt2020towards}'s use of temporal correlations in video data as weak supervision.
We argue the temporal and sequential nature of language can be exploited in a similar manner as well: more precisely, we can consider tokens belonging to the same sequence to share latent factors, and therefore pair up tokens within sequences in a similar way as image pairs in vision.
We next operationalize this idea in our setup (Sec.~\ref{sec:causal_loss}) and then analyze the features obtained by SAEs trained with this method (Sec.~\ref{sec:causal_features}). 
We discuss in Sec.~\ref{sec:causal_analysis} the connection between the inductive biases of the proposed pipeline and the nature of identified features.
We note that the proposed protocol and results should be deemed a proof-of-concept: our goal is to make a stronger connection to prior literature on autoencoder-based approaches to interpretability, hence eliciting (in)abilities of such protocols.

\subsection{Defining Causal Loss}
\label{sec:causal_loss}
%
We propose an additional \textit{causal regularization} term in the loss to incentivize causality. Thus, we now have a loss function given by
\begin{equation*}
\footnotesize
\mathcal{L} = \mathcal{L}_\text{recon} + \alpha\mathcal{L}_\text{sparse} + \beta\mathcal{L}_\text{caus}.    
\end{equation*}

In order to define $\mathcal{L}_\text{caus}$, note that what we want is for interventions on the latents to lead to interpretable changes in the model output. However, we do not want the changes to be arbitrary, i.e., we cannot simply try to maximize the change caused by an ablation---then the incentive for disentanglement is harmed. We therefore introduce weak supervision, motivated by the `match pairing' approach of~\citet{locatello2020weakly, klindt2020towards}.
%
%

Specifically, we try to make the run (forward pass) of the model on a given token $t_1$ similar to its run on a different, albeit related, token $t_2$.
More precisely, given a single token $t_1$, we try to intervene on its latent representation $l_1$ to cause its logit distribution to resemble that of $t_2$. 
%
%
We operationalize this by a simple interpolation technique---we intervene on $l_1$ during the reconstructed run (as defined in Sec.~\ref{sec:exp_causality}) of the model on $t_1$, replacing it with $\lambda \cdot l_1 + (1-\lambda) \cdot l_2$ for some $\lambda \in [0, 1)$. 
To define what to compare the output of the model on this corrupted run to be, we follow our intuition of using the run on $t_2$ as a form of supervision and compare the model output to the interpolation of the outputs of $t_1$ and $t_2$ (Fig.~\ref{fig:causal_loss}). 
This is similar to prior work that aims to improve the robustness of representations and obtain smoother decision boundaries by training on interpolations of inputs~\citep{verma2019manifold}.
Another way to motivate this operationalization is that we would like to force the model to change its output based on specific, surgical changes in the latents that happen as tokens evolve over time.

More formally, let $a \overset{\lambda}{\leftrightarrow} b$ denote the interpolation $\lambda a + (1- \lambda) b$ of vectors $a$ and $b$ by a scalar $\lambda$.
Then we define the causal loss term for a token $t_1$, given a `baseline' token $t_2$, as
\vspace{-0.3em}
\begin{equation*}
\label{eq:def}
\footnotesize
\mathcal{L}_\text{caus}(t_1, t_2) := d\Big(g\left(l_1 \overset{\lambda}{\leftrightarrow} l_2\right), g(l_1) \overset{\lambda}{\leftrightarrow} g(l_2)\Big),
\end{equation*}
where $d$ denotes MSE. 

To pick $t_2$ in the pipeline above, we aim for as minimal an intervention as possible; this is to avoid the latent being completely overwritten, and to ensure that even small, surgical changes in the vicinity of the latent have causal effects. Thus we simply find the token with the latent \textit{nearest} (by $L_2$ distance) to $l_1$. This is done at the sequence level; thus, for each sequence, we pair each token with the one nearest to it in the latent space, and find $\mathcal{L}_\text{caus}$ for each of these pairs. For $\lambda$, we simply sample from the uniform distribution over $[0, 1]$ every iteration---this incentivizes causal effects for a wide range of interventions.

\subsection{Nature of Causal Features}
\label{sec:causal_features}

\begin{figure}
\centering
    \begin{subfigure}{0.5\linewidth}
        \includegraphics[width=\linewidth]{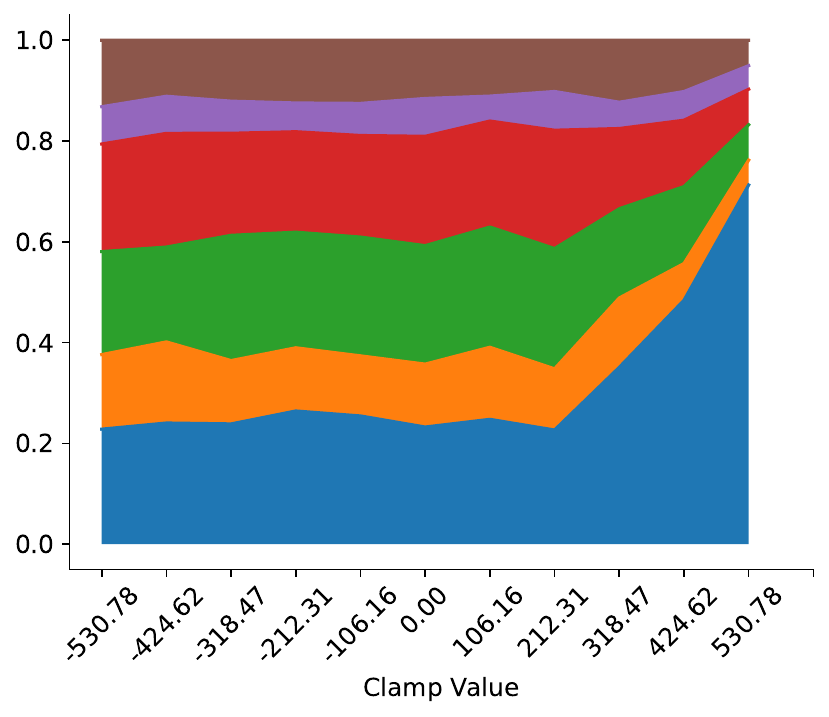}
    \end{subfigure}
    \begin{subfigure}{0.48\linewidth}
        \includegraphics[width=\linewidth]{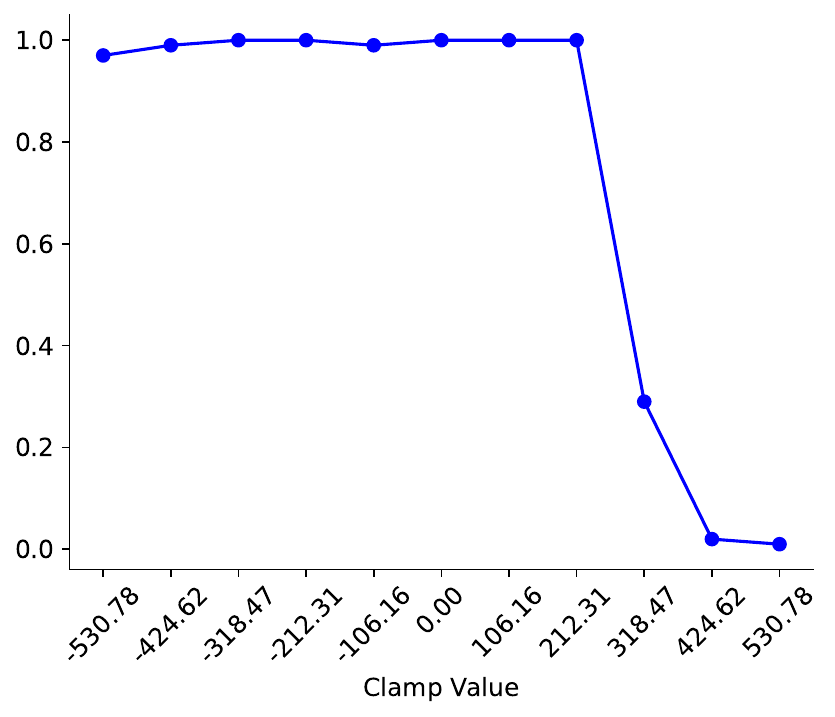}
    \end{subfigure}
\caption{\textbf{Behavior of the English model under interventions.} We intervene on the model, replacing its hidden representations with the SAE's reconstructions, clamping a single latent (in this case, the one corresponding to adjectives) to fixed value. These fixed values are selected at uniform intervals from $[-v_\text{max}, v_\text{max}]$, where $v_\text{max}$ is the maximum value taken by that latent (in line with \citet{templeton2024scaling}). For each value of the clamp (x-axis), we plot the fraction of each part of speech (\textcolor{blue}{nouns}, \textcolor{orange}{pronouns}, \textcolor{green}{adjectives}, \textcolor{red}{verbs}, \textcolor{violet}{adverbs}, and \textcolor{brown}{conjunctions}) in the generated text (\textbf{left}) and the fraction of generations that are grammatical (\textbf{right}). We see that the SAEs trained with causal regularization have a predictable causal function.}
\label{fig:caus_intervention_lcausal}
\end{figure}

As in the SAEs trained without this loss (Sec.~\ref{sec:features}), we identify a number of features correlating with parts of speech---specifically, adjectives, verbs and adverbs. Now, we predict that if these features have a monotonic causal function, they shift the output logit distribution towards the one predicted by the corresponding part of speech. For example, a feature correlating with adjectives should promote the probability of nouns being predicted next, as nouns are the only PoS allowed to come after adjectives (see App.~\ref{sec:grammars_app} for the exact grammar). In the case of verbs, anything may appear as the next token except another verb; thus we expect the probability of verbs to be downweighted by this feature.

Fig.~\ref{fig:caus_intervention_lcausal} presents the effect of these interventions on the PoS distribution across sentences, along with the grammaticality of generations. \textit{The effect is as we hypothesized}. 
We also note that as soon as the distribution significantly shifts away from the uncorrupted distribution, the fraction of grammatical generations drops drastically.
This supports, as pointed out in~\citet{trauble2021disentangled, ahmed2020causalworld}, the difficulty of learning disentangled representations from correlated data (in this case, with respect to PoS distribution and grammaticality).
In fact, recent work~\cite{bhalla2024towards,wu2025axbench} observes that intervention-based methods (in comparison with probing-based methods) suffer from the drawback of damaging models' output quality in terms of coherence, creating a control-capabilities tradeoff.
This indicates that the problem is possibly more general than simple semantic disentanglement, and this is a valuable avenue for future research.
We note, furthermore, that this is another validation of our setup (like the scaling law described in Sec.~\ref{sec:features}), as it demonstrates behaviors also observed in more complex, natural language settings.
More examples of the causal function of these features can be found in App.~\ref{sec:pos_causal}.

\subsection{Inductive Biases in our Pipeline}
\label{sec:causal_analysis}
As we remarked before, \citet{locatello2019challenging} demonstrated that finding latents underlying the data generating process is generally an ill-defined problem with multiple viable solutions.
The authors thus argue that to the extent an approach achieves disentangled features, it is an inductive bias of the training pipeline.
For example, their results show that a good deal of the variance in performance across paradigms (37\%) can be accounted for by the effect of the objective function alone. 
%
%
In this section, therefore, we investigate the inductive biases of our approach. In particular, why do only certain parts of speech (adjectives, verbs and adverbs) have latents correlating with them when our causal loss term is introduced?

\begin{figure}
    \centering
    \includegraphics[width=1.0\linewidth]{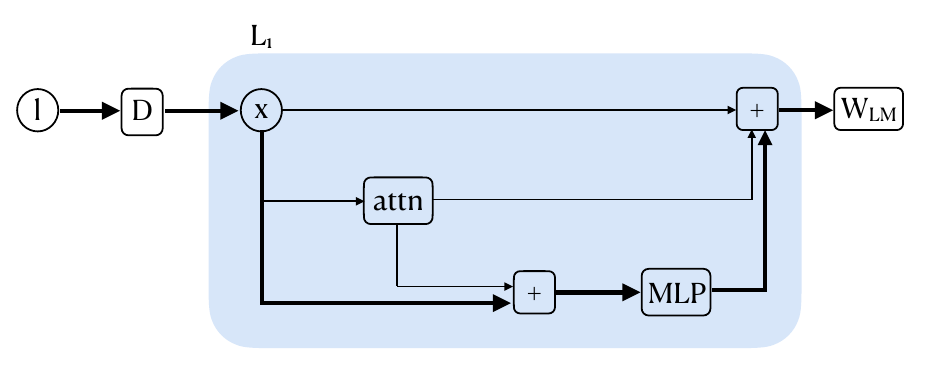}
    \caption{\textbf{The computational graph of the second layer $L_1$ of the transformer model.} Starting from the SAE latent $l$, the decoder $D$ produces a reconstruction of the model activation $x$; the blue box then represents $L_1$, whose output is projected by $W_\text{LM}$ into the logit space. The boldface arrows represent paths through the graph that involve fewer nonlinearities; our causal loss term incentivizes $D$ to write to the subspace of $x$ that is read by the modules in these paths.}
    \label{fig:comp_graph}
\end{figure}

To answer this, we reexamine the causal loss (Sec.~\ref{sec:causal_loss}). As $\overset{\lambda}{\leftrightarrow}$ represents a convex combination for all $\lambda \in [0, 1]$, it effectively incentivizes the function $g$ to be distributive over convex combinations (i.e., over addition and scalar multiplication). In other words, it is an incentive for $g$ to be a \textit{linear} function. This is of course not possible in full generality (else we would not need nonlinear models), but can work for features that have an approximately linear mapping to the output space. 

In particular, consider the computational graph that $g$ represents (Fig.~\ref{fig:comp_graph}). 
Following the framework of \citet{elhage2021mathematical},%
\footnote{This work characterizes the effect of a residual module in a transformer as ``reading from'' (some subspace of) the residual stream, performing some computations, and then writing to (a possibly distinct subspace of) the residual stream. In our model, as shown in Fig.~\ref{fig:comp_graph}, the attention and MLP both function as residual modules.} %
we consider the SAE decoder $D$ as \textit{writing to a subspace} of $x \in \mathbb{R}^d$, which each of $\operatorname{attn}(x)$ and $\operatorname{MLP}(x)$ then \textit{read from}.
Now, note that the attention module involves two softmax operations, and is therefore a highly nonlinear operation. The MLP, by contrast, is a simple linear-GeLU-linear operation, and thus involves only one nonlinearity (which is in fact linear in the range $\mathbb{R}^{+}$). Thus, we claim that the more $D$ writes to the MLP's input subspace, the `more linear' it is; and the more it writes to the attention's input subspace, the `more nonlinear' it is. In other words, the `linearity incentive' of the causal loss manifests itself as a preference for $D$ to have more causal effect on the MLP than on the attention module. For more evidence of this, refer to App.~\ref{sec:linear_app}.

To connect this to the parts of speech we have seen above, we note that it is exactly these parts of speech whose next-token distribution is \textit{static}---adjectives, verbs, and adverbs. In other words, regardless of where they appear in the sequence, the next-token distribution is fixed by the part of speech---i.e., these tokens \textit{do not require attention} for their next-token distribution. For instance, we remark that adjectives can only be followed by nouns---thus the next-token distribution of any adjective token assigns all the probability mass to the nouns in the vocabulary. Verbs and adverbs, too, behave the same way at any position. 

By contrast, the tokens that can follow a noun or pronoun depend on whether or not it is a subject (only conjunctions or verbs) or an object (only conjunctions, adverbs, or \texttt{<EOS>}). Similarly, the tokens that can follow a conjunction depend on whether it connects two noun phrases (only nouns, adjectives, or pronouns) or verb phrases (only verbs).

Putting our subspace intuitions together with the fact that \textit{these} parts of speech do not require attention, we find a plausible answer---the objective creates a signal to promote learning of features that have causal effects on the `more linear' modules (the MLP), while features that affect the attention module are dispreferred. 
This also explains why our causal loss does not find features in the Dyck-2 and Expr models; these languages rely on stacks or counters for generation (which are implemented by attention, as shown by~\citet{bhattamishra2020ability}), rather than vocabulary complexity, and so the attention module carries more of the computational weight. We find further evidence for this in the fact that SAEs trained on these languages \textit{without} causal loss do \textit{not} identify any interpretable features other than those computed by attention.

Overall, in line with prior work on autoencoders-based interpretability~\citep{locatello2019challenging}, our results highlight the importance of recognizing the inductive biases of any SAE paradigm for interpretability. \textit{We expect that any method aiming to integrate causality into feature identification will similarly have inductive biases, and no single method will overcome all limitations of the paradigm.}

\section{Conclusion}
\label{sec:conclusion}
Inspired by studies in vision~\citep{locatello2019challenging, locatello2020weakly}, we propose a minimalistic setup to assess challenges in the use of SAEs for model interpretability. 
We validate our setup by identifying semantically meaningful features and demonstrating the sensitivity of said features' extraction to inductive biases.
We further demonstrate a lack of causality, i.e., interventions on these features do not yield intended effects.
We expect these results to bear out at scale as well, e.g., we find it likely that features identified by SAEs will not always be causally relevant to model computation. 
While previous works, like \citet{marks2024sparse}, have successfully built upon SAE features to identify circuits, we believe our results demonstrate the importance of embedding causality into the feature identification process as a first-class citizen.
As a step towards this integration, we propose a modified objective for training SAEs that exploits the structure of our data, and find that it succeeds in identifying features with predictable causal function. 
We deem these results as a proof-of-concept corroborating our broader arguments, and hypothesize that designing a singular protocol that works well for all tasks and modalities may be difficult.


\section{Limitations}
\label{sec:limitations}
An important caveat for our results is that they come from a simplified, synthetic domain (viz., formal languages); although see App.~\ref{sec:synth} for brief overview of how synthetic pipelines have been helpful to advancing our understanding of deep learning. 
We do believe that the essence of our results is likely to carry over to natural data---in fact, a recent contemporary work by \citet{chaudhary2024evaluating} already corroborates several of our claims.
Nonetheless, it must be kept in mind that significant complications can arise from this domain shift. 
For example, as shown by \citet{lachapelle2023synergies}, full-distribution data in natural language strongly disadvantages the identification of interpretable features, which can be remedied by a restriction to task-specific data. Our formal languages do not show this complication, presumably because the full-distribution variance is of a comparable order to the task-specific variance of natural language.
Furthermore, our proposed causality regularization, while it succeeds in incentivizing the identification of causal directions, strongly prefers a certain kind of feature; i.e., those that form the input to the MLP module. We provide a qualitative explanation for this phenomenon, hoping to inform the application of methods like ours to more real-world settings. We consider overcoming this limitation---perhaps via methods other than regularization---an important future avenue of research.

\section*{Contributions}
ESL conceived the project direction and defined the broader project narrative around disentangled representation learning. ESL wrote code for the data generation process and model training / evaluation, with inputs from AM in expanding to several languages. AM led the feature interpretability analysis, with several inputs from ESL. ESL and AM co-wrote the paper. ESL devised the causal loss idea for Section 4, and AM implemented the method and qualitatively analyzed the results. DSK acted as primary senior advisor, with inputs from MS.

\section*{Acknowledgements}
The authors thank Demba Ba, Sumedh Hindupur, Thomas Fel, Leo Gao, Naomi Saphra, Core Francisco Park, Shashwat Singh and Pratyaksh Gautam for useful conversations during the course of this project. AM was supported by BERI during the course of this project.

\clearpage
\bibliography{custom}

\newpage
\appendix
\onecolumn

\section*{Appendix}

\section{Related Work}
\label{sec:related_app}
Prior work that has influenced the SAE approach to interpretability can be organized into four main directions: disentanglement (App.~\ref{sec:disentanglement}), inductive biases (App.~\ref{sec:inductive}), evaluation and metrics (App.~\ref{sec:metrics}), and the nature of SAE features (App.~\ref{sec:corr}). We also discuss prior work on the use of synthetic data for interpretability (App.~\ref{sec:synth}).

\subsection{Disentanglement and Interpretability}
\label{sec:disentanglement}
\citet{locatello2019challenging}, working in the image domain, carry out a wide-ranging empirical and theoretical study of VAEs on synthetic image data, and prove a non-identifiability result on the learned latent representations. In other words, infinitely many possibilities for disentangled latent representations exist if only a data reconstruction objective is used, and so the disentanglement of the actual representations learned is extremely sensitive to the inductive biases of the autoencoder being used. Thus, no guarantees about the interpretability or task-specific usefulness of the learned representations can be assumed.
\citet{locatello2019challenging} also note that identifiability results are in general not obtainable for the case of a nonlinear data-generating process~\citep{hyvarinen1999nonlinear, hyvarinen2017nonlinear, hyvarinen2019nonlinear}.
Furthermore, \citet{trauble2021disentangled} observe that real-world data can only be generated from highly correlated latents, possibly with a complex causal relationship. They prove also that disentangled representations do not represent an optimum in this case, and so entangled representations are learned. However, they also note that supervision can be leveraged to achieve true disentanglement – auxiliary data linking priors to observations can be used for weak supervision during training to resolve latent correlations.
Later works outlined settings in which identifiability results can in fact be proven. For example, \citet{lachapelle2023synergies} propose a bi-level optimization problem, where the representations are optimized for reconstruction as well as performance on downstream tasks via sparse classifiers, and \citet{squires2023linear} demonstrate how to achieve disentanglement with interventional data (data from observations with individual latents ablated).

Empirically, however, SAEs have been used to obtain many insights on the functioning of neural models. For instance, \citet{demircan2024sparseautoencodersrevealtemporal} use SAE features to draw connections between in-context learning (ICL) is implemented and temporal-difference (TD) learning, a reinforcement learning algorithm; and \citet{lan2024sparseautoencodersrevealuniversal} show that various LLMs have similar representational spaces, through matching the dictionaries learnt by SAE decompositions of these spaces.

\subsection{Scaling Laws and Inductive Biases in SAEs}
\label{sec:inductive}
\citet{locatello2019challenging} highlight the importance of drawing attention to the inductive biases of autoencoders while using them to achieve disentanglement. They show that the objective function and random seed together are responsible for roughly 80\% of the performance of VAE encoders, demonstrating the lack of robustness in the method.
Many works have also obtained empirical results on the relationship between the dictionary size (\textit{i.e.}, the latent size of the autoencoder) and the features learned. For example, \citet{sharkey2024takingout} (working with numerical data) note that recovering the ground-truth features requires a dictionary size of 1--8$\times$ the size of the input, and that if sparsity is enforced by $L_1$-regularization, then larger dictionary sizes need larger penalties. Other studies have found that ``dead'' features (\textit{i.e.}, features that don't activate on any sample) begin to occur from a dictionary size of about 4x~\citep{cunningham2023sparse}, that a single feature in small SAEs ``splits'' into several features (whose union represents the former feature) in larger SAEs~\citep{makelov2024towards, bricken2023towards}, and that several features are simply the same token in various contexts, like a physics ``the'' and a mathematics ``the''~\citep{bricken2023towards}.

\subsection{Metrics}
\label{sec:metrics}
Many aspects of SAEs have been identified as important for evaluation, and many metrics exist for each of these. Mainly, they can be classified along two axes: interpretability vs. disentanglement, and supervised vs. unsupervised.
Supervised metrics require some ground-truth dictionary of features to evaluate against, which is generally assumed to be human-interpretable. Therefore, interpretability and disentanglement are tied together in these metrics. For example, BetaVAE uses the accuracy of a classifier trained to predict ground-truth features from learned ones~\citep{locatello2019challenging, sepliarskaia2019not}; consistency and restrictiveness measure the sensitivity of ground-truth features to learned ones~\citep{shu2019weakly}; and maximum mean cosine similarity (MMCS) maps the two sets of features using the cosine similarity~\citep{sharkey2024takingout}.
However, when no ground-truth is available, a feature set may be disentangled but not interpretable, or vice versa. Thus, unsupervised metrics evaluate interpretability and disentanglement separately. Examples of metrics for interpretability are controllability, which evaluates how `easy' it is to control the model output by intervening on a feature set~\citep{makelov2024towards}, and next logit attribution, where the causal role of the feature in the model's final logit output is examined~\citep{bricken2023towards}.
Notably, \citet{makelov2024towards} find that SAEs trained on task-specific data (IOI in their case study) learn meaningful directions, while those trained on full data perform on par with those that have random directions kept frozen through training.

\subsection{Correlational and Causal Features}
\label{sec:corr}
A number of studies demonstrate the causal effects of SAE features. For example, \citet{bricken2023towards} show that the features representing Arabic-language text (in a one-layer transformer) can be clamped to a high value, increasing the probability of generating Arabic text. \citet{templeton2024scaling} scale up this work to analyze Claude's representations, identifying a feature representing the Golden Gate Bridge in San Francisco, with a causal effect on how prominent this monument is in the model's output. 
Notably, \citet{marks2024sparse} use SAE directions to identify circuits in models (across layers) responsible for specific tasks, like subject-verb agreement across relative clauses.
However, as shown in our results, the bulk of features have no causal effects on the model computation (using standard SAE protocols). We claim similar results can be easily demonstrated in natural settings as well---for instance, \citet{chaudhary2024evaluatingopensourcesparseautoencoders} report results qualitatively similar to ours in a restricted natural language setting as well.

\citet{braun2024identifyingfunctionallyimportantfeatures} also propose a modified training objective, called end-to-end(e2e) SAEs, to ensure that SAEs learn about the model rather than the dataset. They show that e2e SAEs obtain an improvement over standard SAEs in terms of reconstruction and interpretability.

\subsection{Synthetic Data in Interpretability}
\label{sec:synth}
Synthetic data from controlled, well-understood domains has long been used to investigate the functioning and capabilities of language models. This data takes various forms---\citet{li2022emergent} train a model to predict valid moves from a textualized representation of the board game Othello, and show that the model learns an internal representation of the board state that can be intervened on to change its predictions, and \citet{li2024chain} use arithmetic tasks, like modular addition, to characterize the capabilities of chain-of-thought prompting over standard decoding.

Formal languages have also supplied synthetic data, particularly to understand the expressivity of various sequence architectures. We have discussed in the main paper (Sec.~\ref{sec:features}) the results of~\citet{bhattamishra2020ability}, who use the languages of Dyck-1, Shuffle-Dyck and boolean expressions to establish the role of attention in generalization across sequence lengths, and contextualize the set of languages recognizable by transformers within the Chomsky hierarchy~\citep{sipser1996introduction}.

In the domain of interpretability, \citet{cagnetta2024deep} use a class of CFGs, called random hierarchy models (RHMs), to identify how transformers learn compositional structures and quantify the data required to learn a hierarchical task. \citet{allenzhu2024physicslanguagemodels1} also use probes to identify the mechanism by which transformers learn to recognize CFGs, identifying syntactic structures in hidden states and information-passing functions in attention patterns. In this connection, the results of \citet{wen2024transformers} are also relevant---they use the Dyck language to show that the attention module may be `nearly randomized' without affecting the performance of the model. This provides important context for interpretability methods oriented towards isolated submodules of transformers, or `myopic' methods.

Generative paradigms other than context-free grammars have also been used to investigate models---for instance, \citet{lubana2024percolationmodelemergenceanalyzing} use context-sensitive languages to investigate and model the phenomenon of \textit{emergent} capabilities, or sudden improvement in certain tasks at certain points during training.

\section{Formal Grammars}
\label{sec:grammars_app}
We use, as mentioned in Sec.~\ref{sec:experiments}, three formal grammars on which we train transformer-based language models. The exact specification of these grammars, in terms of a context-free grammar (CFG), is presented below.
Note that the actual data generation process is defined by a probabilistic CFG, which requires probabilities to be assigned to each of the productions of a nonterminal. We omit these probabilities for legibility here, but readers can refer to our  (\url{https://github.com/Abhinav271828/pcfg-sae-causal-arr-oct24}) for details.

\subsection{Dyck-2}
Given $n$ types of brackets (that is, $2n$ symbols consisting of $n$ opening and the matching $n$ closing brackets), the Dyck-$n$ consists of all the valid sequences of brackets. 
Algorithmically, a string belonging to Dyck-$n$ can be parsed by maintaining a stack of opening brackets, and popping the topmost one when the corresponding closing bracket is encountered. If a closing bracket that does not match the topmost opening bracket is encountered, the string is rejected.
The production rules that express the generation of strings from Dyck-2, then, are as follows.
\begin{align*}
    S &\to S \, S \, \mid B_1 \, \mid B_2 \\
    B_1 &\to \texttt{(} \, S \, \texttt{)} \mid \texttt{( )} \\
    B_2 &\to \textbf{[} \, S \, \texttt{]} \mid \texttt{[ ]}
\end{align*}

\subsection{Expr}
\label{sec:counter}
The Expr language is the set of prefix arithmetic expressions, in which operands are single digits from 0 to 9, and operators may be unary, binary or ternary. There are three operators of each type.
Note that it is possible to view the vocabulary as being organized according to \textit{arity}, or number of arguments needed. Thus digits are symbols of arity 0; unary operators of arity 1; binary operators of arity 2; and ternary operators of arity 3.

\begin{algorithm*}
\caption{Identify points in the sequence where exactly one more expression is expected, without maintaining an explicit counter.}\label{alg:almost_done}
\begin{algorithmic}
\Require Tokens $t_1, \dots, t_n$ \\
$i = 1$
\While{$i \leq n$}
\If{$t_i$ is a unary operator}
    \State mark $t_i$
    \State $i \gets i + 1$
\ElsIf{$t_i$ is a binary operator}
    \State $i \gets$ the position of the last token of the first operand of $t_i$
    \State mark $t_i$
\ElsIf{$_i$ is a ternary operator}
    \State $i \gets$ the position of the last token of the second operand of $t_i$
    \State mark $t_i$
\Else
    \State $i \gets i + 1$
\EndIf
\EndWhile
\end{algorithmic}
\end{algorithm*}

Since the syntax is prefix, there is no need to define precedence for unambiguous parsing. For parsing, it is sufficient to maintain a counter that keeps track of how many more expressions are needed to complete the sequence – this counter starts at 1 (as the whole sequence, which is pending, represents one expression), and is incremented by $n-1$ when we encounter a token of arity $n$. Thus, for instance, if 3 more expressions are needed to complete the sequence and a binary operator is encountered, we now need 4 more expressions – two to complete the binary operator, and two more to satisfy the original three. Parsing succeeds if this value reaches 0 at the end of the string, and fails if it becomes negative at any point during the parse.

The production rules for this language are as follows.
\begin{align*}
    S &\to O \mid D \\
    D &\to 0 \mid 1 \mid 2 \mid 3 \mid 4 \mid 5 \mid 6 \mid 7 \mid 8 \mid 9 \\
    O &\to U \, S \mid B \, S \, S \mid T \, S \, S \, S \\
    U &\to \text{un}_1 \mid \text{un}_2 \mid \text{un}_3 \\
    B &\to \text{bin}_1 \mid \text{bin}_2 \mid \text{bin}_3 \\
    T &\to \text{tern}_1 \mid \text{tern}_2 \mid \text{tern}_3
\end{align*}

As we note in Sec.~\ref{sec:experiments}, SAEs trained on Expr models find a feature that activates on tokens where exactly one more expression is needed to complete the sequence, \textit{i.e.}, where the counter above has a value of 1. We consider it reasonable to assume the implicit computation of such a counter based on the existence of this feature, since the algorithm required to identify these tokens (Algorithm~\ref{alg:almost_done}) is much more convoluted if we forbid explicitly computing this counter.

\subsection{English}
We define a simple fragment of English, intended to capture major parts of speech and bridge the gap between parsing languages like Dyck-2 and Expr above, and natural language parsing. We retain the two most common sentence constructions, but ignore more complicated syntactic features like agreement and relative clauses, and morphological features, like conjugations and declensions.
This grammar can be parsed using any standard CFG parsing algorithm, like Earley or CKY parsing~\citep{sipser1996introduction}.

The rules for the grammar are as follows.
\begin{align*}
    S &\to \text{NP} \, \text{VP} \\
    \text{NP} &\to \text{Pro} \mid \text{N} \mid \text{NP} \; \text{Conj} \; \text{NP} \mid \text{Adj} \, \text{N} \\
    \text{VP} &\to \text{V} \mid \text{V} \, \text{NP} \mid \text{VP} \; \text{Conj} \; \text{VP} \mid \text{VP} \, \text{Adv}
\end{align*}

Each part of speech – nouns (N), verbs (V), pronouns (Pro), conjunctions (Conj), adjectives (Adj) and adverbs (Adv) – have ten tokens each. We omit the production rules listing these for brevity, but refer readers to the  (\url{https://github.com/Abhinav271828/pcfg-sae-causal-arr-oct24}) for details.

\begin{table*}
\centering
\begin{tabular}{p{5cm}lll}
\hline
\textbf{Explanation} & \textbf{Language} & \textbf{Settings} & \textbf{Correlation} \\
\hline
One more expression required to complete the sequence. & Expr & (2, $k = 16$) & 0.972 \\
Last token. & Expr & (8, $\alpha = 10^{-3}$) & 0.984 \\
Verbs. & English & (8, $k = 128$) & 0.964 \\
Adjectives. & English & (8, $k = 128$) & 0.985 \\
Adverbs. & English & (8, $k = 128$) & 0.999 \\
Conjunctions. & English & (8, $k = 128$) & 0.932 \\
Empty stack of type 0. & Dyck & (8, $k = 128$) & 0.925 \\
Empty stack of type 1. & Dyck & (8, $k = 128$) & 0.814 \\
Stack depth 11 or more. & Dyck & (8, $k = 128$) & 0.924 \\
All brackets at depth 0, and the first opening and all closing brackets at depth 1. & Dyck & (8, $k = 128$) & 0.912 \\
\hline
\end{tabular}
\caption{\textbf{Features identified by SAEs.} We give a description of each feature, with the language it is found in, and the correlation (Pearson coefficient) between the activations and the explanation. All the SAEs are trained without \texttt{pre\_bias} or normalization; we therefore identify them by their expansion factor and regularization factor ($\alpha$ in the case of $L_1$- and $k$ in the case of top-$k$ regularization).}
\label{table:features}
\end{table*}

\section{Features}
\label{sec:features_app}
We describe here the methodology of finding features in SAEs, and list several other features we were able to identify.

\subsection{Feature Identification Pipeline}
\subsection{Experimental Protocols: Feature Identification}

We avoid the automated interpretability paradigm outlined in~\citet{bills2023language}, in order to avoid concerns of LLMs being unfamiliar with the specific formal grammars we use.
Instead, in line with~\citet{bricken2023towards}, we begin with \textit{manual inspection} of each feature, and hypothesize potential explanations for its semantics using positive and negative samples.
As mentioned in the same work, this must also be informed by the reconstruction capabilities of the SAE---we therefore restrict our analysis to SAEs that are able to achieve high reconstruction accuracy (defined in~\ref{sec:sensitivity}.
Following~\citet{makelov2024towards}, we compare the activations of the latents with the predictions of the explanations. We use the correlation between the two as a score for the quality of the explanation.
\citet{marks2024sparse} use a manual interpretability pipeline similar to ours, with the added information of the latent's causal role (through ablations).
However, our aim is primarily to evaluate the quality of purely correlational explanations, which many feature identification pipelines (both manual and automated) rely upon.
The causal nature of the explanations is explored in Sec.~\ref{sec:lack_causality}

The manual inspection is based on an \textit{a priori} notion of the latent generative factors inherent in each formal language, which we briefly outline.
In Dyck-2, we expect some computation to be based on the depth of the current token (\textit{i.e.,} the number of unclosed brackets preceding it).
In Expr, we expect a counter that keeps track of the number of expressions yet to be generated.
In English, we expect features corresponding to each part of speech in the grammar.
These are all variables \textit{both necessary and sufficient} for the data-generation process. We note, however, that these are not the only possibilities---each language may have several equivalent reformulations, but we believe these are the most intuitive and suitable for the autoregressive decoding paradigm.
We discuss in more detail the exact formulation of the grammars and their relation to these generative factors in App.~\ref{sec:grammars_app}.

\subsection{Identified Features}
Results are shown in Tab.~\ref{table:features}.
Note that all the SAEs listed here are trained without \texttt{pre\_bias} and without normalization. Thus, the hyperparameters specified in column 3 are the expansion factor (ratio between hidden size and input size) and, according to the regularization method, either the $L_1$ coefficient or $k$.
For each feature, we note the hyperparameter settings of the SAE, our hypothesized explanation, and the correlation (Pearson coefficient) between the feature's activation and the explanation.

In the Expr model, the SAEs identify features relating to a counter variable (based on the parsing process described in Sec.~\ref{sec:counter}) and position. Of the former kind, we find a feature that activates exactly when there is one more expression left to complete the sequence; as an example of the latter, there are features that activate only on the last token of a sequence.

In the model trained on a fragment of English, we find features representing several parts of speech – adjectives, verbs, adverbs, and conjunctions. Each of these activate to varying degrees on tokens belonging to the respective part of speech.

In the Dyck-2 model, as in the case of Expr, we see features that correspond to an intuitive stack-based parsing process. In Sec.~\ref{sec:results}, we present a feature that `matches' corresponding opening and closing trends. There are also features that threshold the depth of (number of unclosed brackets that appear before) a token, \textit{i.e.}, they activate when the depth is above a certain value (11 in this case). We also find features that identify a combination of depth and whether a bracket is opening or closing.

\section{Causality of Features}
\label{sec:causality_app}

We present here a detailed outline of our experimental protocols in the feature identification and the causality experiments. We also examine other features for causal effects, particularly the counter feature for Expr (described in Sec.~\ref{sec:results}), and part-of-speech features for English. In the latter case, we examine features identified by ordinary SAEs and SAEs with causal loss in Sec.~\ref{sec:pos_noncausal} and Sec.\ref{sec:pos_causal} respectively. We refer the reader to our  for causal experiments (\url{https://github.com/Abhinav271828/pcfg-sae-causal-arr-oct24}) for the implementation of these protocols.

\subsection{Experimental Protocols}
As explained in Sec.~\ref{sec:exp_causality}, in order to examine the causal effects of a latent identified by an SAE, we rely on interventions that replace this latent by some fixed value, and continue the computation of the larger model.
In other words, consider a computational graph of the language model (see Fig.~\ref{fig:causal_loss}). The node corresponding to the layer that is being examined, \textit{i.e.}, the first hidden layer, is replaced by three nodes:
\begin{enumerate}[leftmargin=15pt, itemsep=3pt, topsep=2pt, parsep=2pt, partopsep=2pt, label=\roman*.]
\item the actual representations generated by the previous layer, which form the input to the SAE;
\item the hidden layer \textit{of the SAE}, which consists of the features identified by the SAE; and
\item the output, or the reconstruction, of the SAE.
\end{enumerate}
The last node feeds back into the larger model, where the original input should have been used. Thus, we effectively replace the model activations with SAE reconstructions of those activations. We then intervene on node (ii) above – we select a single element in the SAE representation, set it to our value, and recompute the modified reconstruction, which is then used in the rest of the LM's forward pass.

The exact intervention that we carry out is defined in terms of the maximum value $v_\text{max}$ that the feature attains across a sample of 1280 sequences. We then select the values for the intervention by spacing 10 intervals across the range $[-v_\text{max}, v_\text{max}]$, \textit{i.e.}, the intervention values are
$$v_i = -v_\text{max} + \frac{i}{10} \cdot 2v_\text{max}, \forall i \in \{0, 1, \dots, 10\}.$$

The principle of using $v_\text{max}$ as the baseline for our interventions is in line with \citet{templeton2024scaling}. Note that the above leads to 11 possible values for the interventions.

\subsection{English: Parts of Speech (without causal loss)}
\label{sec:pos_noncausal}
The results of interventions on various features are shown in Fig.~\ref{fig:interventions_app}. For each part of speech that we show, we follow the same process as in the case of adjectives (Fig.~\ref{fig:intervention}):
\begin{itemize}[leftmargin=12pt, itemsep=3pt, topsep=2pt, parsep=2pt, partopsep=2pt]
\item determine the maximum value $v_\text{max}$ taken by a certain latent;
\item pick clamp values at uniform intervals in the range $[-v_\text{max}, v_\text{max}]$; and
\item examine (i) the distribution of parts of speech and (ii) the grammaticality of the generations after clamping the latent to each value.
\end{itemize}

We observe that, similar to the adjectives feature in the main paper (see Fig.~\ref{fig:intervention}), the features for other parts of speech do not show causal effects either.

\subsection{English: Parts of Speech (with causal loss)}
\label{sec:pos_causal}
\begin{figure*}
\captionsetup[subfigure]{labelformat=empty}
    \centering
    \begin{subfigure}{0.23\textwidth}
    \includegraphics[width=0.9\linewidth]{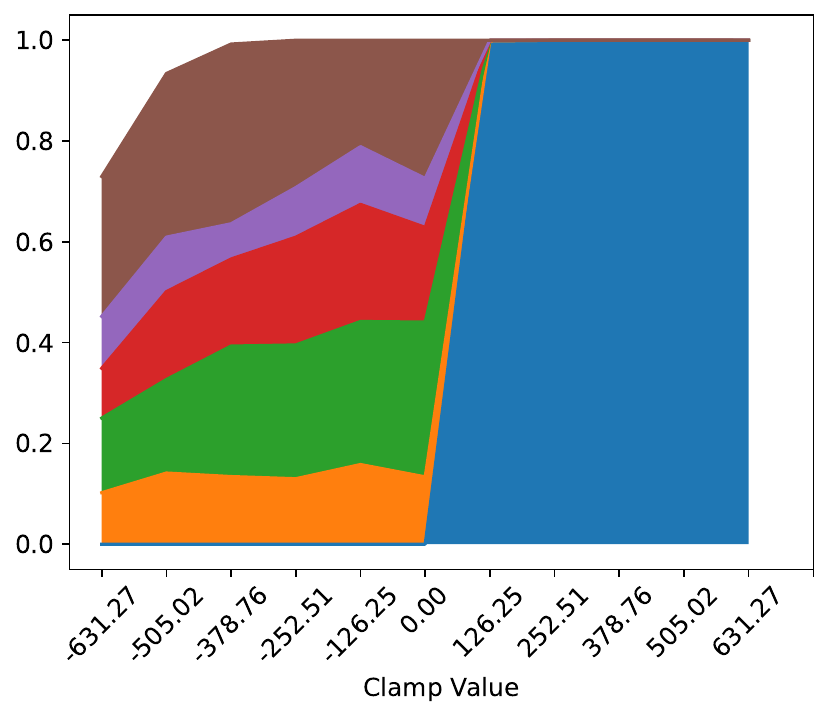}
    \caption{}
    \end{subfigure}
    ~
    \begin{subfigure}{0.23\textwidth}
        \includegraphics[width=0.9\linewidth]{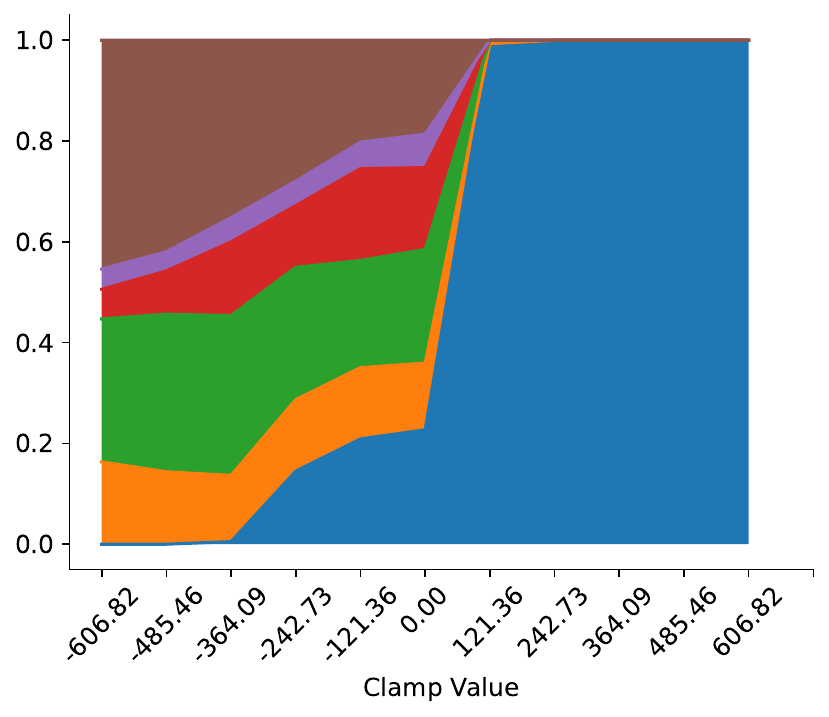}
        \caption{}
    \end{subfigure}
    ~
    \begin{subfigure}{0.23\textwidth}
        \includegraphics[width=0.9\linewidth]{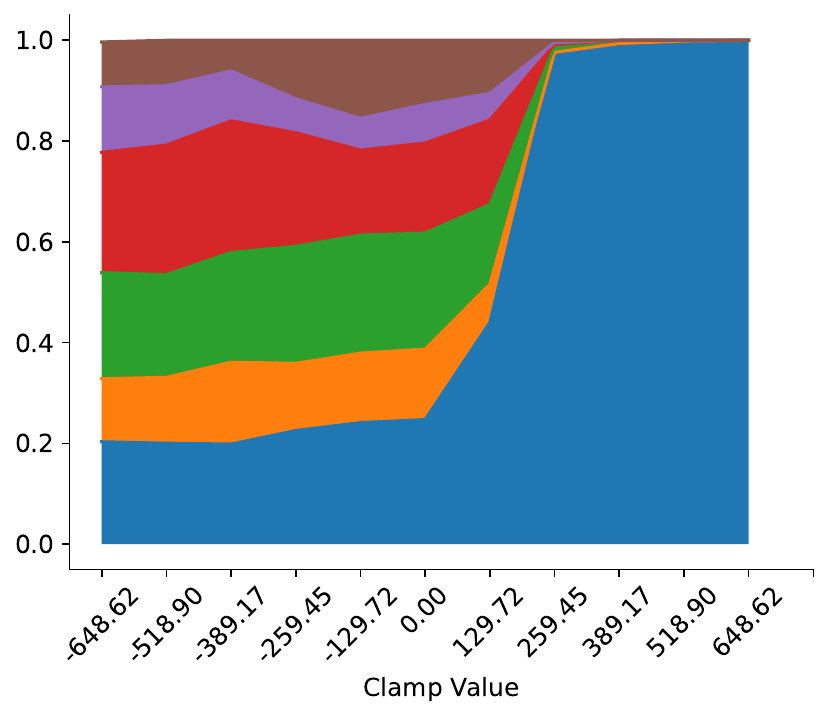}
        \caption{}
    \end{subfigure}
    ~
    \begin{subfigure}{0.23\textwidth}
        \includegraphics[width=0.9\linewidth]{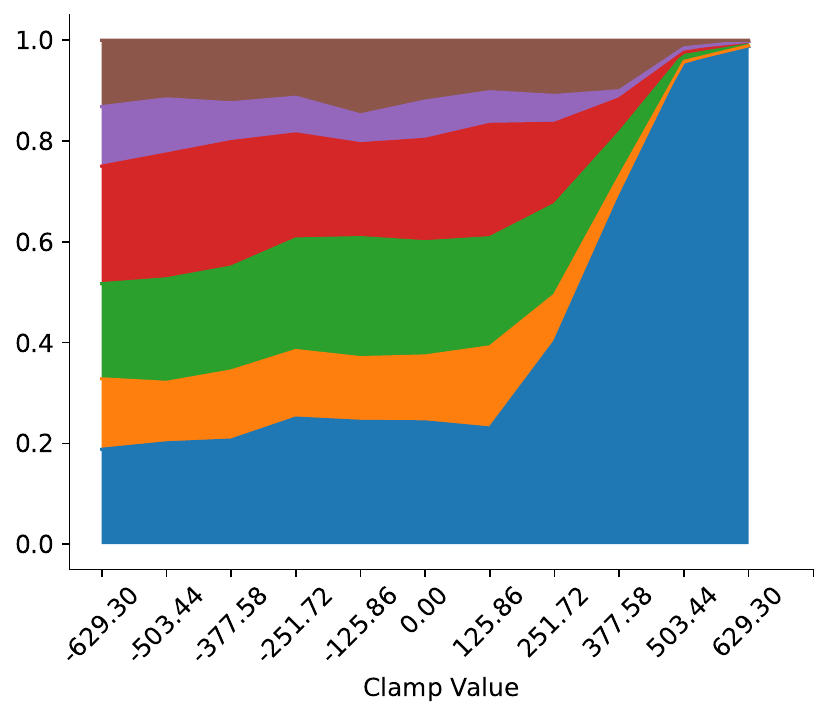}
        \caption{}
    \end{subfigure}
    \begin{subfigure}{0.23\textwidth}
    \includegraphics[width=0.9\linewidth]{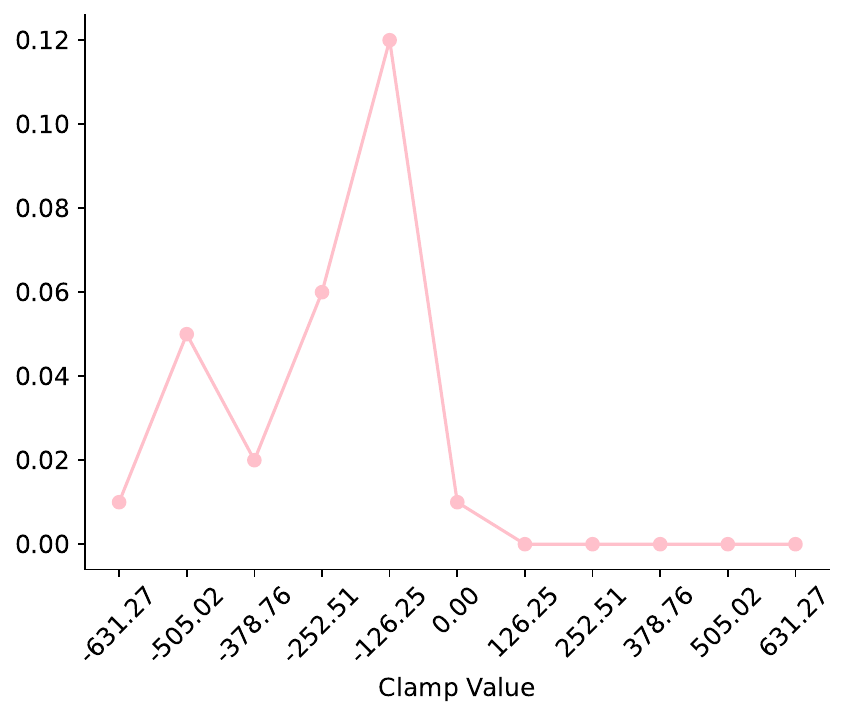}
    \caption{(a) 15/15}
    \end{subfigure}
    ~
    \begin{subfigure}{0.23\textwidth}
        \includegraphics[width=0.9\linewidth]{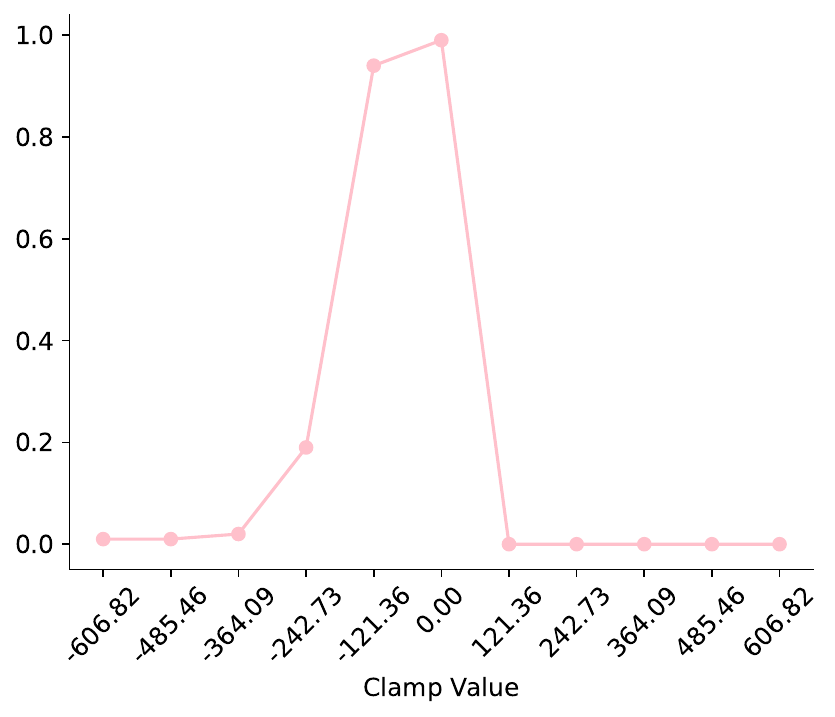}
        \caption{(b) 8/15}
    \end{subfigure}
    ~
    \begin{subfigure}{0.23\textwidth}
        \includegraphics[width=0.9\linewidth]{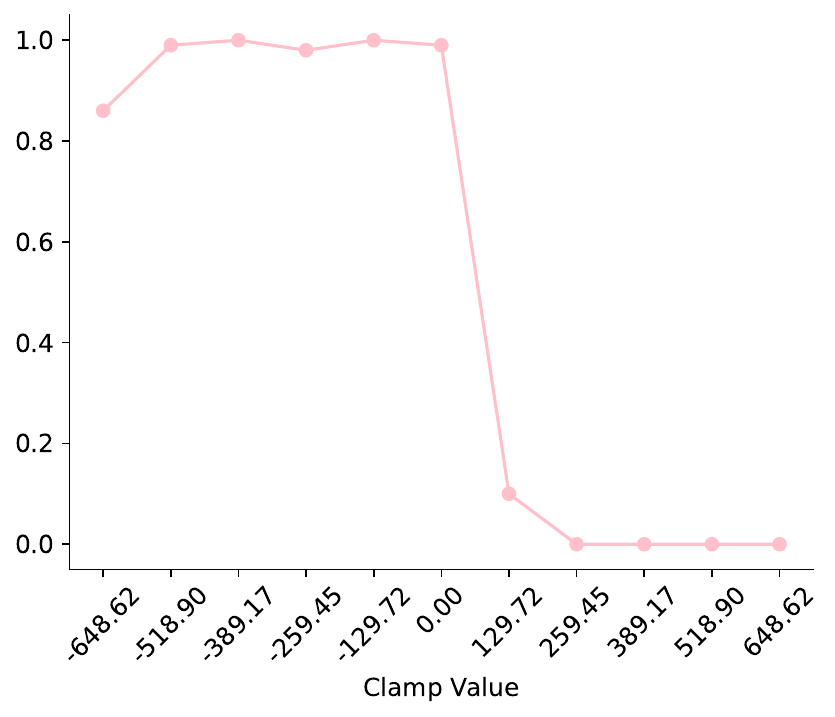}
        \caption{(c) 4/15}
    \end{subfigure}
    ~
    \begin{subfigure}{0.23\textwidth}
        \includegraphics[width=0.9\linewidth]{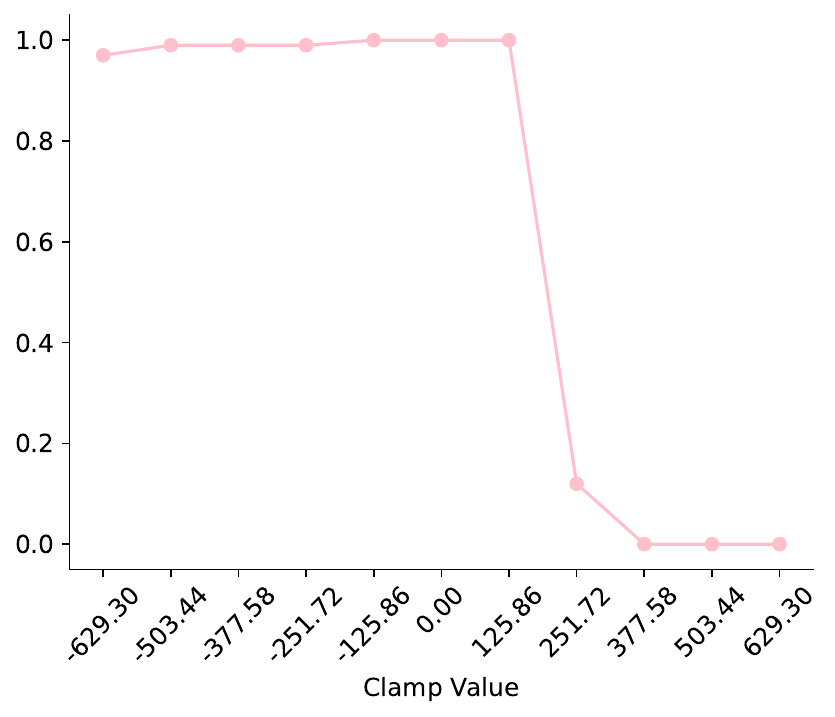}
        \caption{(d) 2/15}
    \end{subfigure}

    \begin{subfigure}{0.23\textwidth}
    \includegraphics[width=0.9\linewidth]{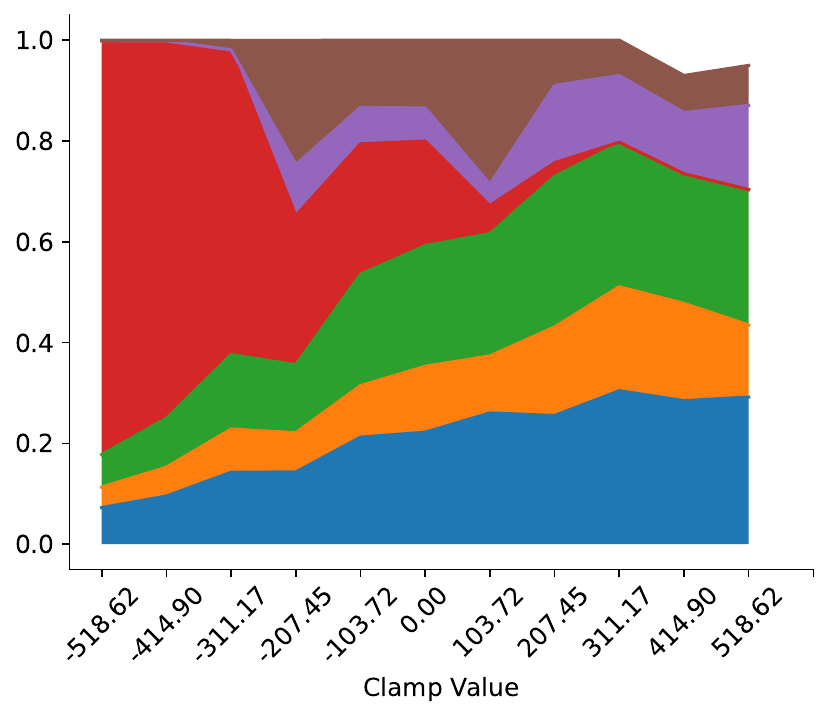}
    \caption{}
    \end{subfigure}
    ~
    \begin{subfigure}{0.23\textwidth}
        \includegraphics[width=0.9\linewidth]{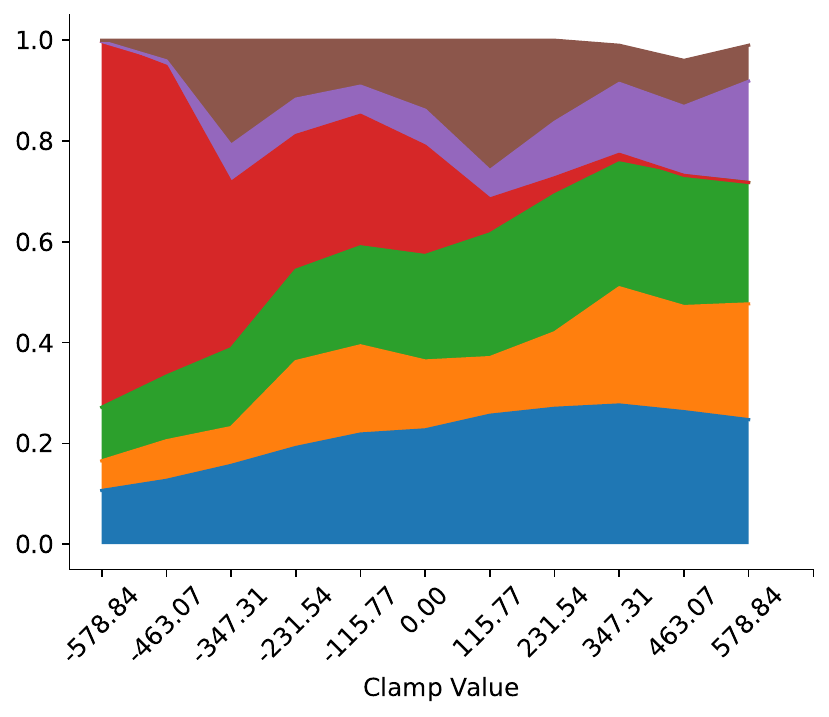}
        \caption{}
    \end{subfigure}
    ~
    \begin{subfigure}{0.23\textwidth}
        \includegraphics[width=0.9\linewidth]{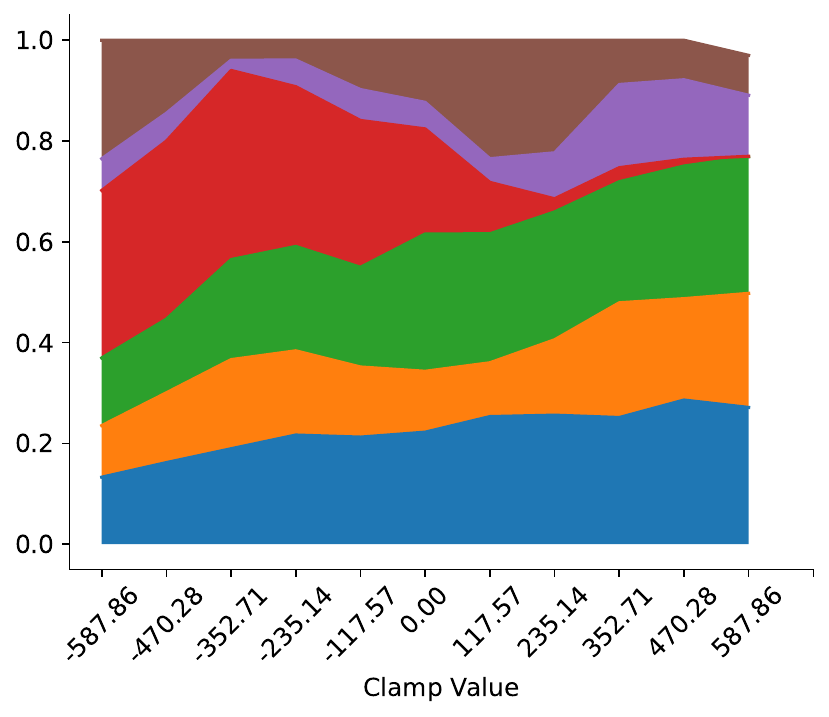}
        \caption{}
    \end{subfigure}
    ~
    \begin{subfigure}{0.23\textwidth}
        \includegraphics[width=0.9\linewidth]{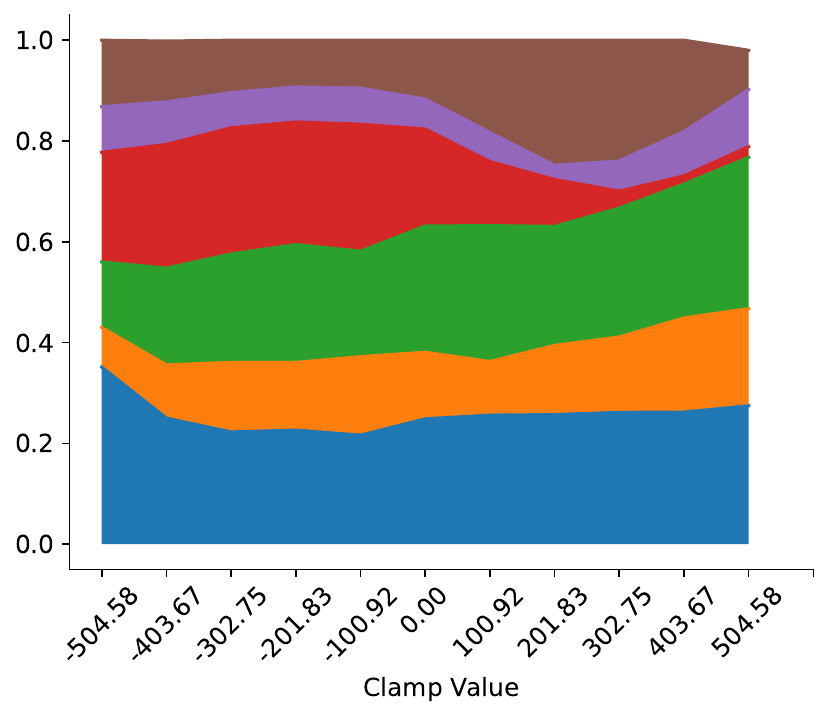}
        \caption{}
    \end{subfigure}
    \begin{subfigure}{0.23\textwidth}
    \includegraphics[width=0.9\linewidth]{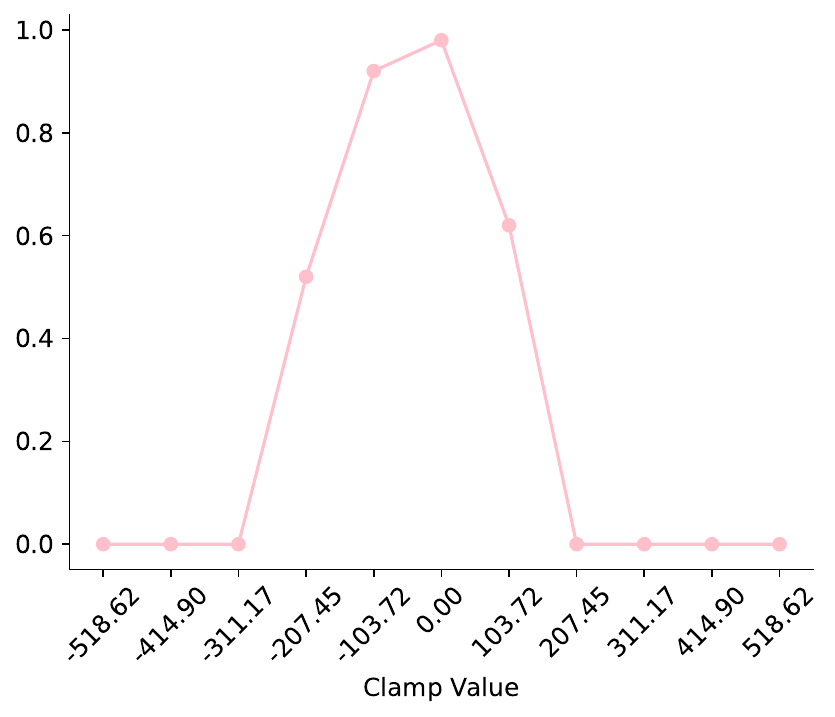}
    \caption{(e) 5/5}
    \end{subfigure}
    ~
    \begin{subfigure}{0.23\textwidth}
        \includegraphics[width=0.9\linewidth]{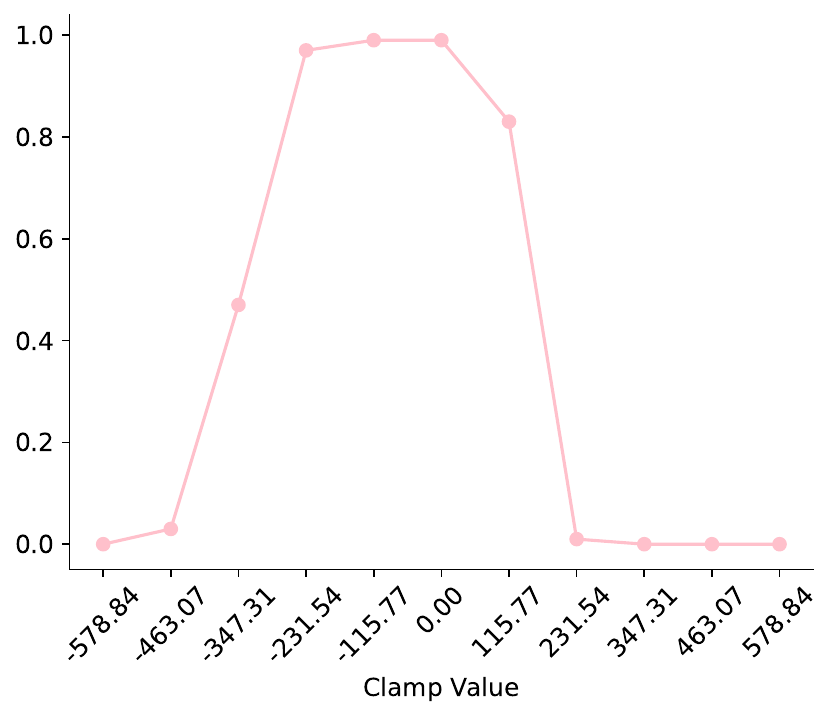}
        \caption{(f) 4/5}
    \end{subfigure}
    ~
    \begin{subfigure}{0.23\textwidth}
        \includegraphics[width=0.9\linewidth]{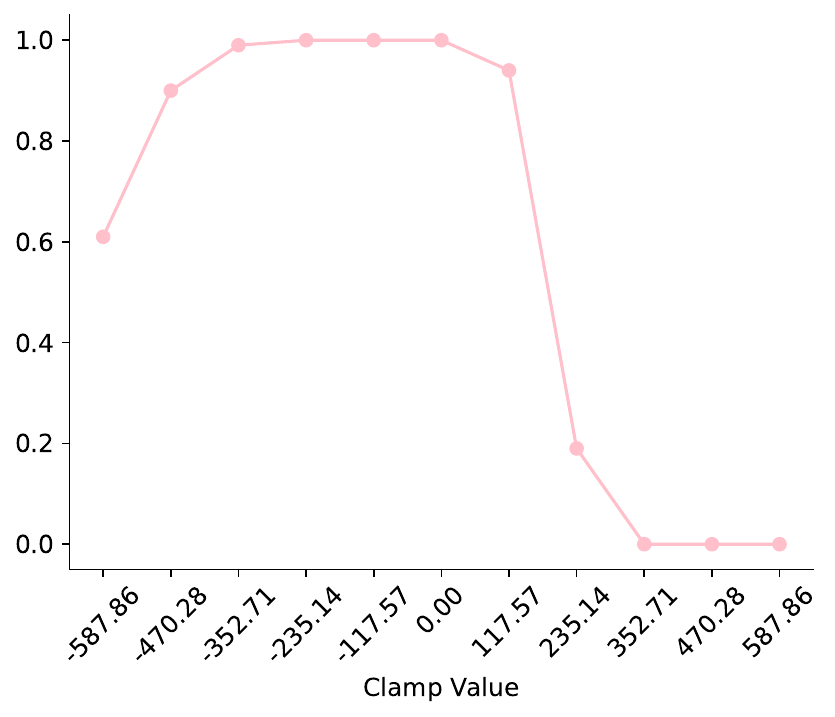}
        \caption{(g) 3/5}
    \end{subfigure}
    ~
    \begin{subfigure}{0.23\textwidth}
        \includegraphics[width=0.9\linewidth]{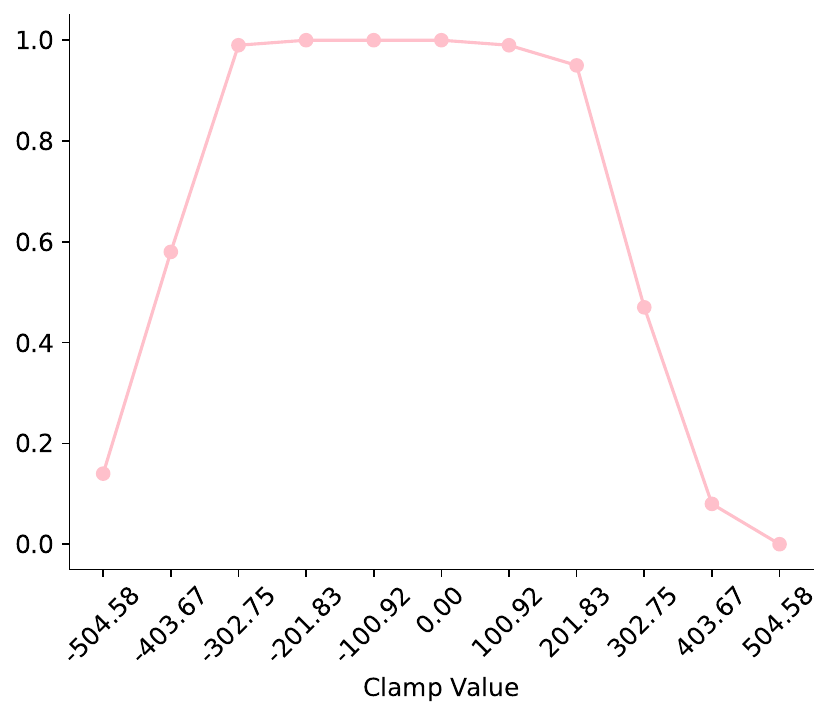}
        \caption{(h) 2/5}
    \end{subfigure}

    \begin{subfigure}{0.23\textwidth}
    \includegraphics[width=0.9\linewidth]{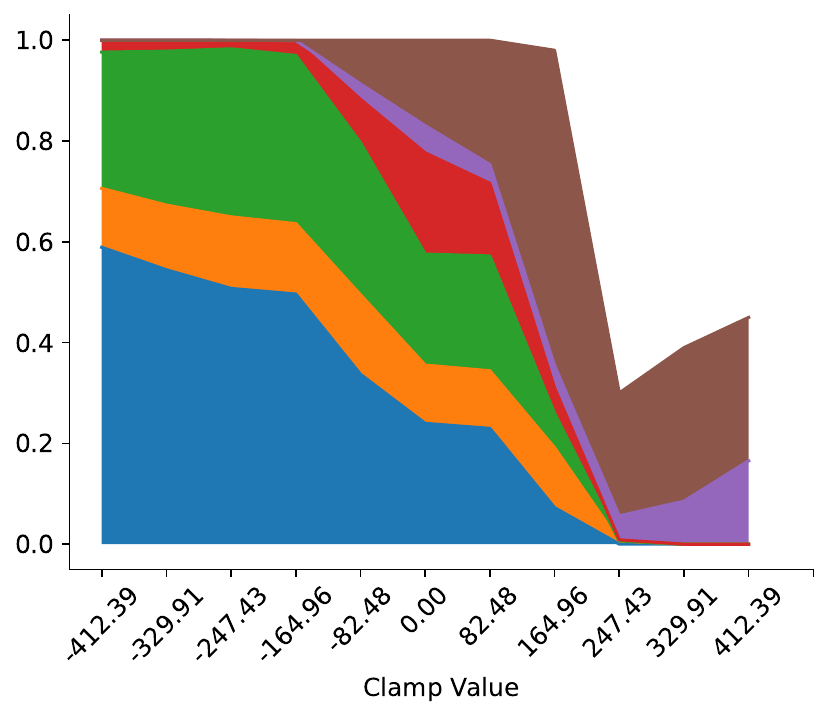}
    \caption{}
    \end{subfigure}
    ~
    \begin{subfigure}{0.23\textwidth}
        \includegraphics[width=0.9\linewidth]{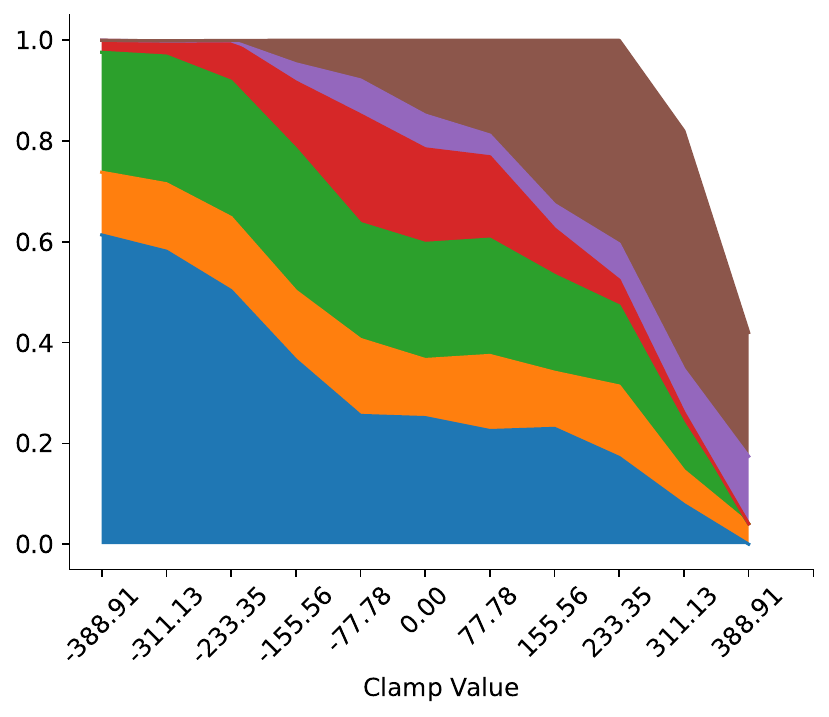}
        \caption{}
    \end{subfigure}
    ~
    \begin{subfigure}{0.23\textwidth}
        \includegraphics[width=0.9\linewidth]{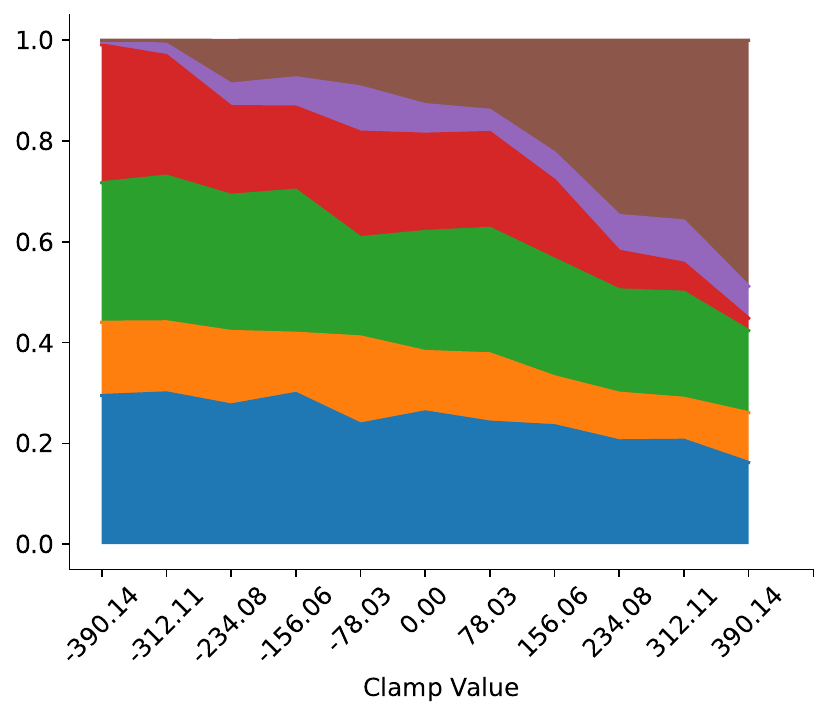}
        \caption{}
    \end{subfigure}
    ~
    \begin{subfigure}{0.23\textwidth}
        \includegraphics[width=0.9\linewidth]{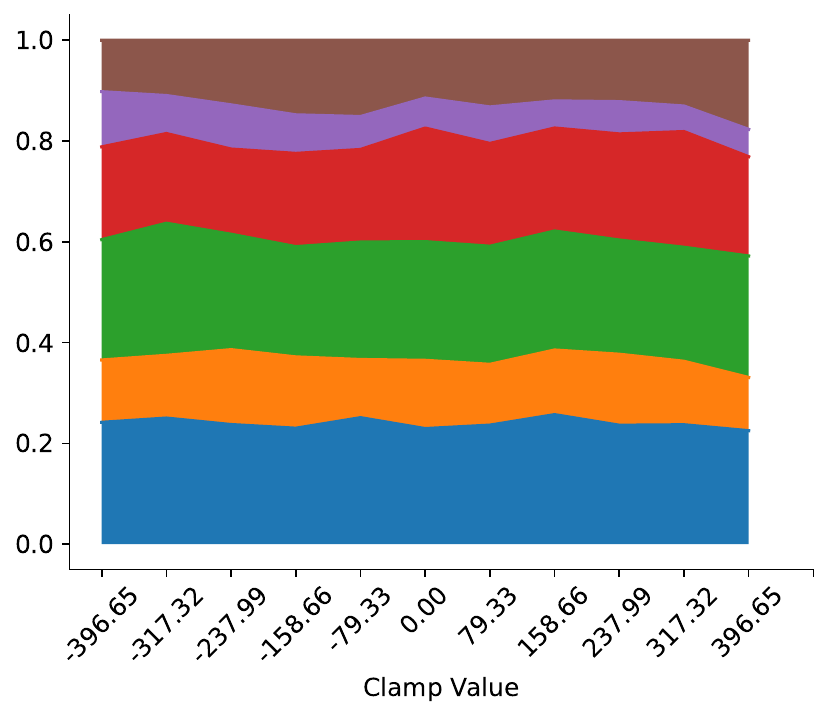}
        \caption{}
    \end{subfigure}
    \begin{subfigure}{0.23\textwidth}
    \includegraphics[width=0.9\linewidth]{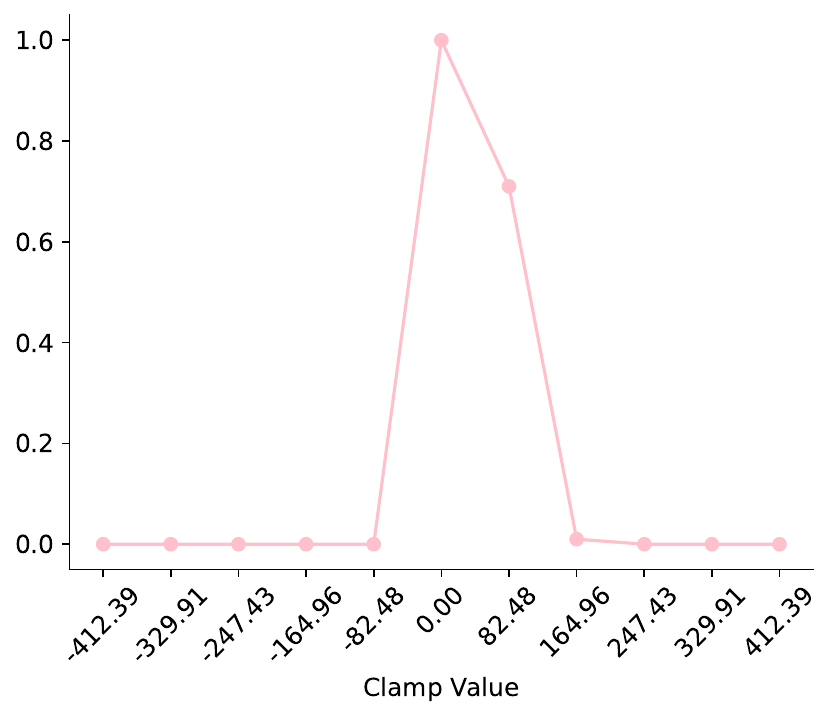}
    \caption{(i) 13/13}
    \end{subfigure}
    ~
    \begin{subfigure}{0.23\textwidth}
        \includegraphics[width=0.9\linewidth]{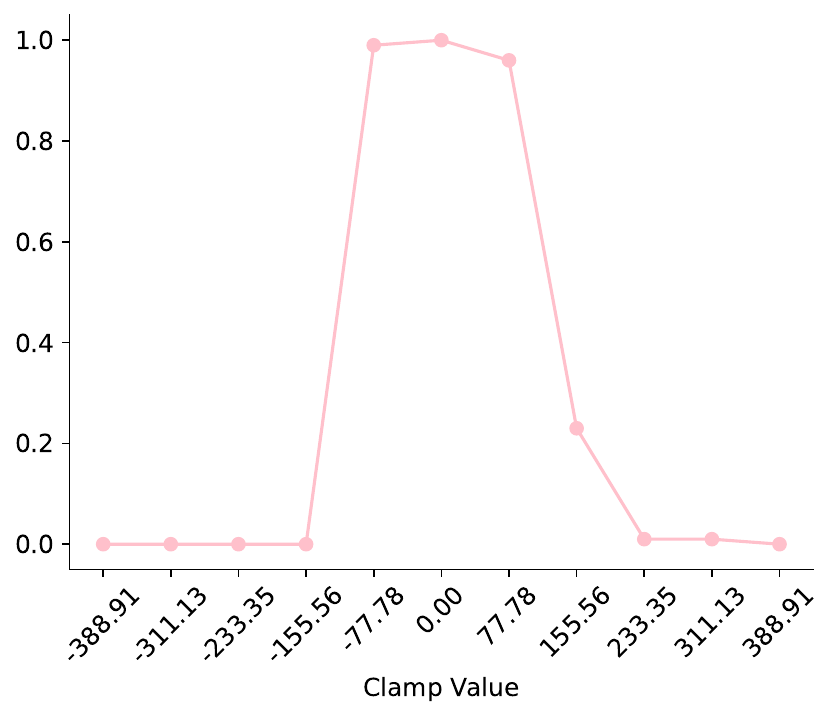}
        \caption{(j) 7/13}
    \end{subfigure}
    ~
    \begin{subfigure}{0.23\textwidth}
        \includegraphics[width=0.9\linewidth]{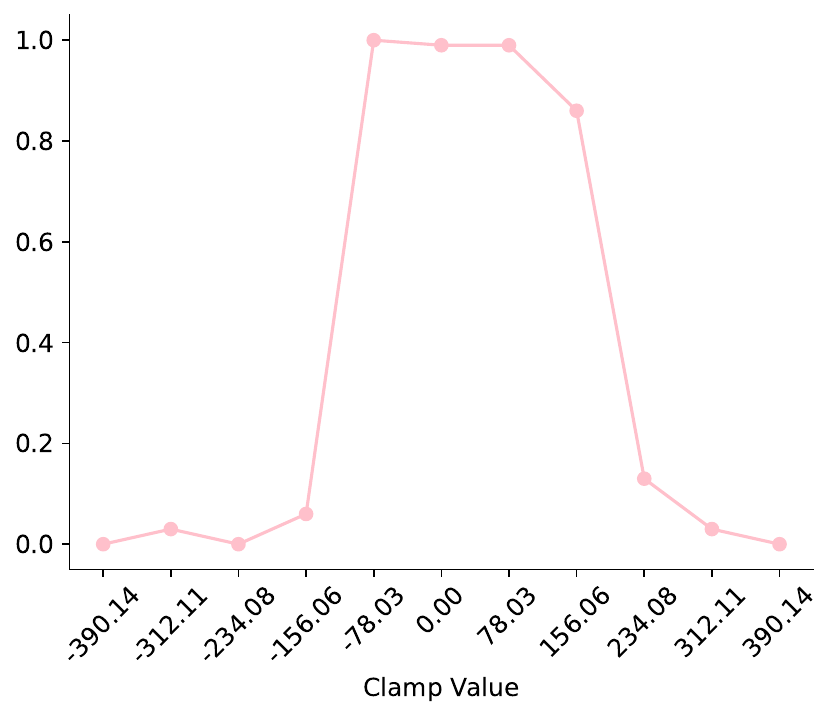}
        \caption{(k) 4/13}
    \end{subfigure}
    ~
    \begin{subfigure}{0.23\textwidth}
        \includegraphics[width=0.9\linewidth]{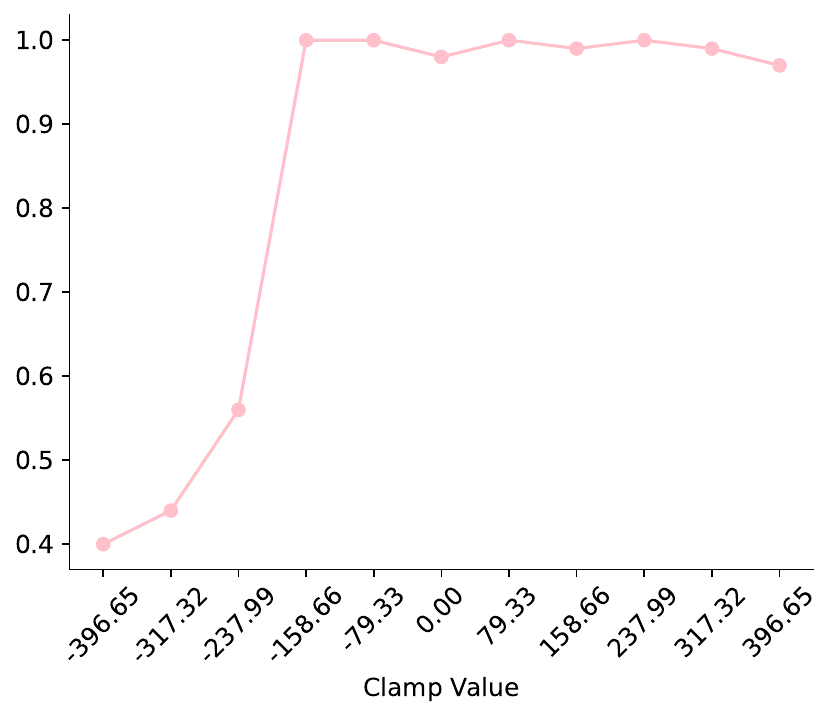}
        \caption{(l) 2/13}
    \end{subfigure}
    \caption{\textbf{Causal effects of \textbf{adjectives} ((a)-(d)), \textbf{verbs} ((e)-(h)), and \textbf{adverbs} ((i)-(l)).} The caption of each image describes the number of latents being intervened on. At each clamp value (x-axis), we plot the distribution of \textcolor{blue}{nouns}, \textcolor{orange}{pronouns}, \textcolor{green}{adjectives}, \textcolor{red}{verbs}, \textcolor{violet}{adverbs}, and \textcolor{brown}{conjunctions} (top) and the percentage of grammatical generations (bottom).}
\label{fig:causal_effects}
\end{figure*}

As mentioned in Sec.~\ref{sec:causality}, our SAEs trained with causal loss demonstrate features correlating with adjectives, verbs, and adverbs. We proceed in essentially the same manner as Sec.~\ref{sec:pos_noncausal} above. As our models, however, identify several features for each part of speech, we show the causal effects of subsets of these feature sets of various sizes, in order to demonstrate the causal effect of the entire set. For instance, if the model has 15 features corresponding to adjectives, we demonstrate causal effects of 15, 8, 4, and 2 features, each of which is obtained by random sampling. We intervene on a set of features in the same way as a single one – by clamping all of the latents to the same value, and resuming the run of the model. The effects of the interventions are shown in Fig.~\ref{fig:causal_effects}.

\begin{table*}
\centering
\begin{tabular}{lll}
    \hline
\textbf{Applied Corruption} & \textbf{Divergence from Clean Run} & \textbf{Divergence from Corrupted Run} \\
    \hline
$\Delta x$ & $10^{-3}$ & $1.37$ \\
$\Delta \operatorname{attn}(x)$ & $9.1 \cdot 10^{-5}$ & $1.38$ \\
$\Delta \operatorname{MLP}(x + \operatorname{attn}(x))$ & $1.29$ & $6.5 \cdot 10^{-3}$ \\
$\Delta_\text{attn} \operatorname{MLP}(x + \operatorname{attn}(x))$ & $1.44$ & $0.04$ \\
$\Delta_x \operatorname{MLP}(x + \operatorname{attn}(x))$ & $4.8 \cdot 10^{-3}$ & $1.53$ \\
    \hline
\end{tabular}
\caption{\textbf{KL divergence between partially corrupted runs, clean runs, and corrupted runs.}}
\label{table:lin_ev}
\end{table*}

Following the logic of Sec.~\ref{sec:causal_features}, we make the following predictions for each part of speech:
\begin{itemize}
    \item \textbf{Adjectives.} As noted in Sec.~\ref{sec:causal_loss}, nouns are the only tokens that can come after adjectives. Therefore we expect the feature correlating with adjectives, if it has a causal effect, to upweight the probability of nouns being predicted.
    \item \textbf{Verbs.} We have also seen in Sec.~\ref{sec:causal_loss} that any part of speech can follow a verb, except another verb. Thus, we expect the probability of verbs to be reduced by high values of verb-correlated features.
    \item \textbf{Adverbs.} The grammar (App.~\ref{sec:grammars_app}) shows that adverbs can only be followed by conjunctions, more adverbs, or the \texttt{<EOS>} token. We then expect the former two types of tokens to be upweighted, and the other parts of speech dispreferred.
\end{itemize}

We observe that, in contrast to the features examined in Sec.~\ref{sec:pos_noncausal} above, these features show clear and predictable causal effects.

\section{Evidence for Linearity Incentive}
\label{sec:linear_app}
Here, we present further evidence for our intuition that the causal loss incentivizes the SAE features to preferentially affect the MLP.

First, we examine the definition of the function $L_1$ (the second layer of the model). Ignoring LayerNorm and dropout, this function is simply
$$L_1(x) = x + \operatorname{attn}(x) + \operatorname{MLP}\left(x + \operatorname{attn}(x)\right).$$

Thus, when the input $x$ is corrupted to, say $x + \Delta x$, each of these terms lends a corresponding corrupted term the output of $L_1$ (and therefore of $g$).

Now, we examine the effect of adding each of these difference terms (separately) to $L_1(x)$. If, as we hypothesize, the MLP is the most significantly affected module, then we expect that
\begin{align}
    L_1(x) &\approx L_1(x) + \Delta x \\
    L_1(x) &\approx L_1(x) + \Delta \operatorname{attn}(x) \\
    L_1(x + \Delta x) &\approx L_1(x) \nonumber \\
        &+ \Delta \operatorname{MLP}(x + \operatorname{attn}(x));
\end{align}
in other words, the effect of the corruption should be replicated by corrupting \textit{only} the MLP output; and corrupting either of the other two terms should have no effect.

Correspondingly, we can carry this analysis to the input of the MLP module. Here, we expect that the difference in the MLP module's output can be achieved by corrupting only $x$, and not $\operatorname{attn}(x)$ at all. To make this precise, we define
\begin{align*}
    \Delta_x \operatorname{MLP}(x &+ \operatorname{attn}(x)) \\
        &= \operatorname{MLP}(x + \Delta x + \operatorname{attn}(x)) \\
    \Delta_\text{attn} \operatorname{MLP}(x &+ \operatorname{attn}(x)) \\
        &= \operatorname{MLP}(x + \operatorname{attn}(x + \Delta x)).
\end{align*}

Then we expect, in line with the previous case, that
\begin{align}
    L_1(x) &\approx L_1(x) \nonumber \\
        &+ \Delta_\text{attn} \operatorname{MLP}(x + \operatorname{attn}(x)) \\
    L_1(x + \Delta x) &\approx L_1(x) \nonumber \\
    &+ \Delta_x \operatorname{MLP}(x + \operatorname{attn}(x));
\end{align}
in other words, the corruption of the attention module does not affect the output of the MLP module. \\
Note that in this setting, the corruption is only applied to the input of the MLP model; the outside $\operatorname{attn}(x)$ term remains as is.

We provide in Tab.~\ref{table:lin_ev} the KL divergence between expressions (1), (2), (3), (4) and (5) above, and the distributions corresponding to $L_1(x)$ and $L_1(x + \Delta x)$. The results are as we expect from the qualitative explanation in Sec.~\ref{sec:causality}. Here, the corruption consists of clamping a feature correlating with adjectives to its maximum value.

\begin{table*}
\centering
\begin{tabular}{llll}
\hline
\textbf{Figure} & \textbf{Regularization} & \texttt{pre\_bias} & \textbf{Normalization} \\
\hline
Fig.~\ref{fig:runset_1} & $L_1$   & False & none \\
Fig.~\ref{fig:runset_2} & top-$k$ & False & none \\
Fig.~\ref{fig:runset_3} & $L_1$   & True  & input, decoder \\
Fig.~\ref{fig:runset_4} & top-$k$ & True  & input, decoder \\
Fig.~\ref{fig:runset_6} & $L_1$   & False & input \\
Fig.~\ref{fig:runset_8} & top-$k$ & False & input \\
Fig.~\ref{fig:runset_7} & $L_1$   & False & input, reconstruction \\
Fig.~\ref{fig:runset_9} & top-$k$ & False & input, reconstruction \\
\hline
\end{tabular}
\caption{\textbf{Hyperparameter Settings and Corresponding Figures.}}
\label{table:runsets}

\end{table*}

It is interesting to note here that we were unable to identify interpretable features in the case of causal SAEs trained on Dyck and Expr models. We explain this, in line with the above, by observing that generation in these languages necessarily requires long-distance memory, in the form of a stack or a counter, which is a function taken up by attention in a transformer model~\citep{bhattamishra2020ability}. Thus the MLP plays much less of a role in these languages (given the lack of diversity in the vocabulary), and attention is much more important. Since our causal loss avoids identifying features that attention relies on, this explains the lack of features in these two languages.

\section{Power Law in Reconstruction Loss}
\label{sec:power_law}
\begin{figure*}
  \centering
\begin{subfigure}[t]{0.45\textwidth}
  \includegraphics[width=1.0\linewidth]{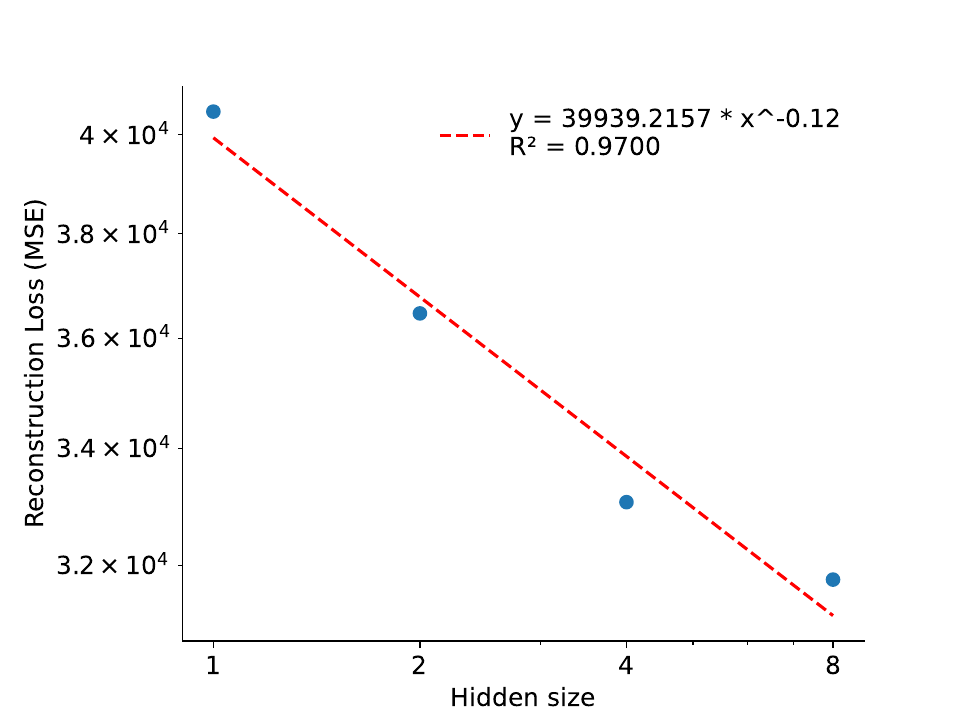}
  \caption{}
\end{subfigure}
\hspace{0.05\textwidth}
\begin{subfigure}[t]{0.45\textwidth}
  \includegraphics[width=1.0\linewidth]{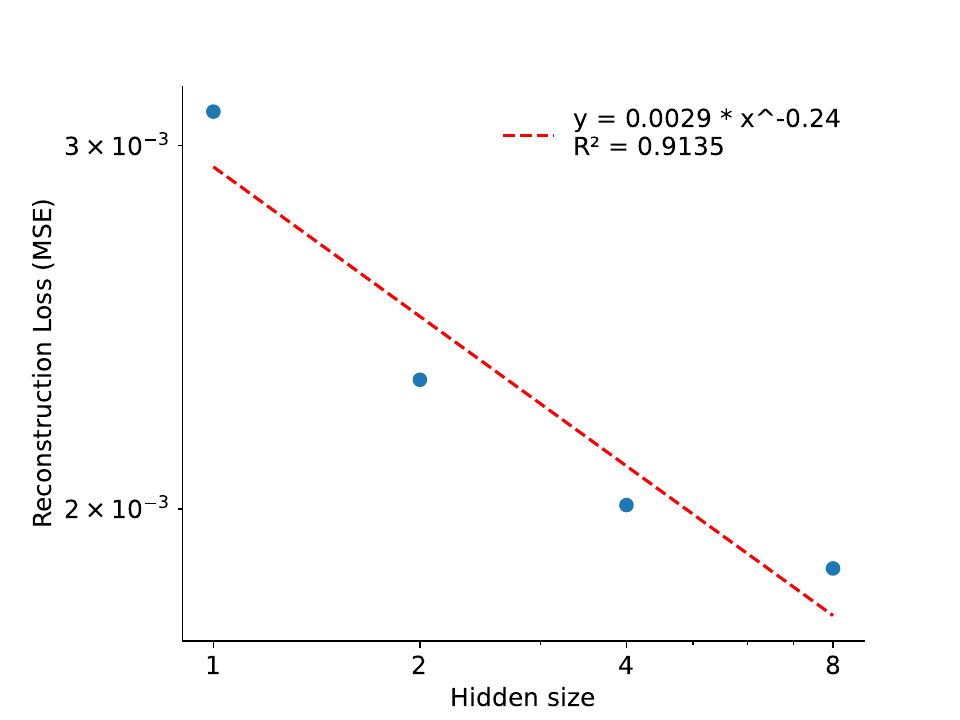}
  \caption{}
\end{subfigure}
\vspace{0.5cm}
\begin{subfigure}[t]{0.45\textwidth}
    \includegraphics[width=1.0\textwidth]{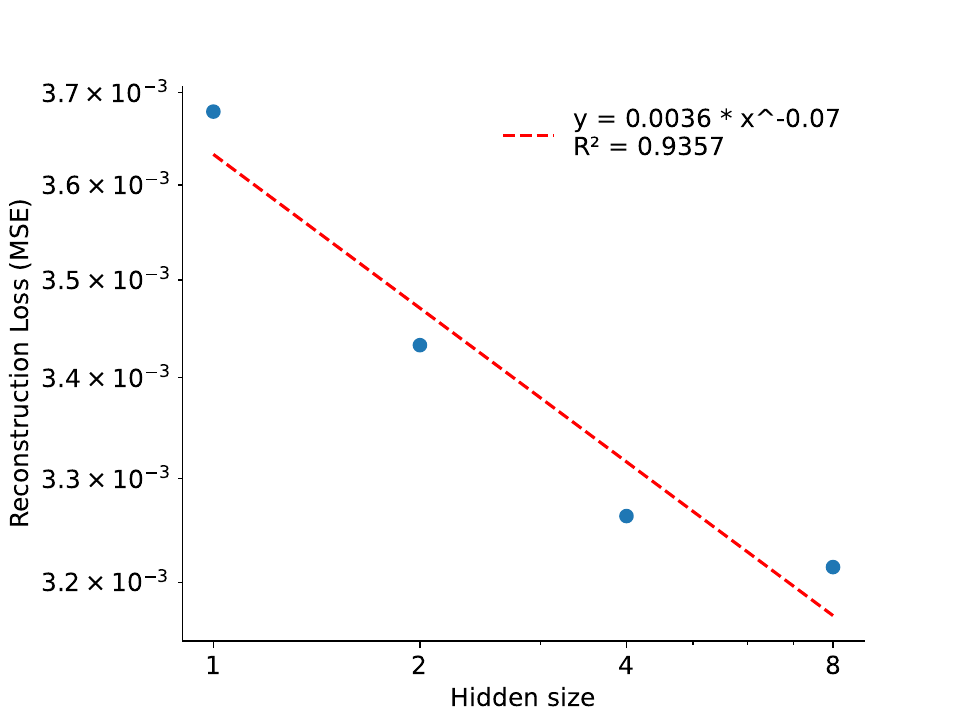}
    \caption{}
\end{subfigure}
\hspace{0.05\textwidth}
\begin{subfigure}[t]{0.45\textwidth}
    \includegraphics[width=1.0\textwidth]{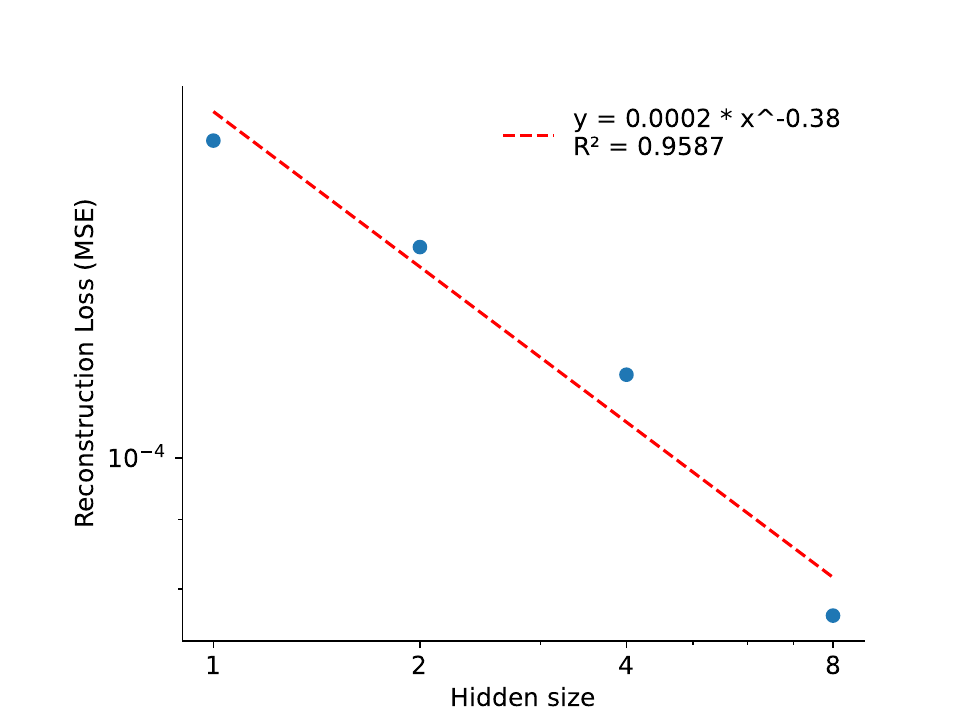}
    \caption{}
\end{subfigure}
\caption{\textbf{Scaling law connecting hidden size to reconstruction loss.} The power law we identify for each top-$k$ regularized set of autoencoders. The four run sets are \textbf{(a)} without \texttt{pre\_bias} or normalization; \textbf{(b)} with \texttt{pre\_bias} and normalized inputs and decoder directions; \textbf{(c)} without \texttt{pre\_bias} and normalized inputs; and \textbf{(d)} without \texttt{pre\_bias} and normalized inputs and reconstructions.}
\label{fig:power_law_all}
\end{figure*}

We noted in Sec.~\ref{sec:results} that the top-$k$ regularized SAEs reveal a power law in their reconstruction losses. The trends for various runs are shown in Fig.~\ref{fig:power_law_all}.

\section{Training Details}
For both our Transformer and SAE training, we use an online data generation process, that randomly generates a sequence from a given PCFG at every iteration. We use two different datasets for the training and validation in both cases.

\begin{figure*}
    \centering
    \begin{subfigure}[t]{0.45\textwidth}
    \includegraphics[width=1.0\linewidth]{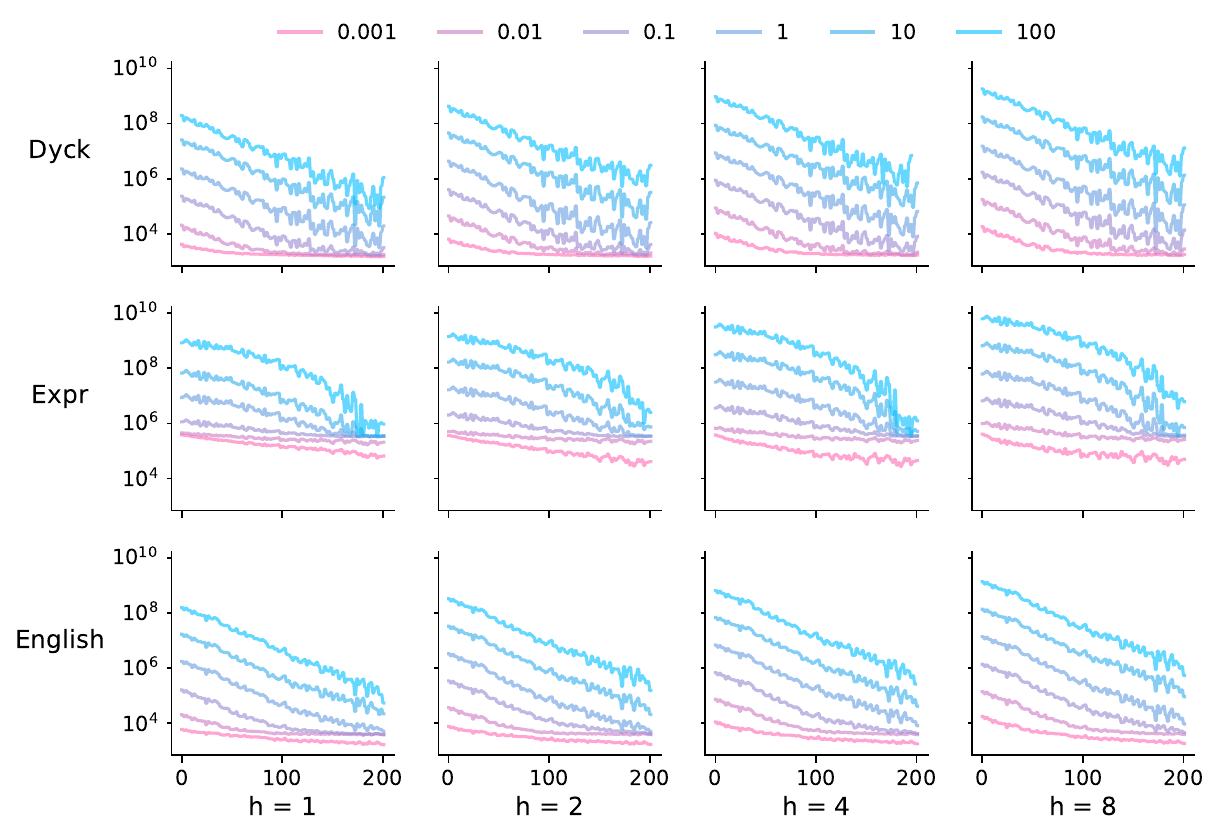}
    \caption{Train Loss (sum of reconstruction and regularization losses)}
    \end{subfigure}
    ~
    \vspace{5mm}
    \begin{subfigure}[t]{0.45\textwidth}
    \includegraphics[width=1.0\linewidth]{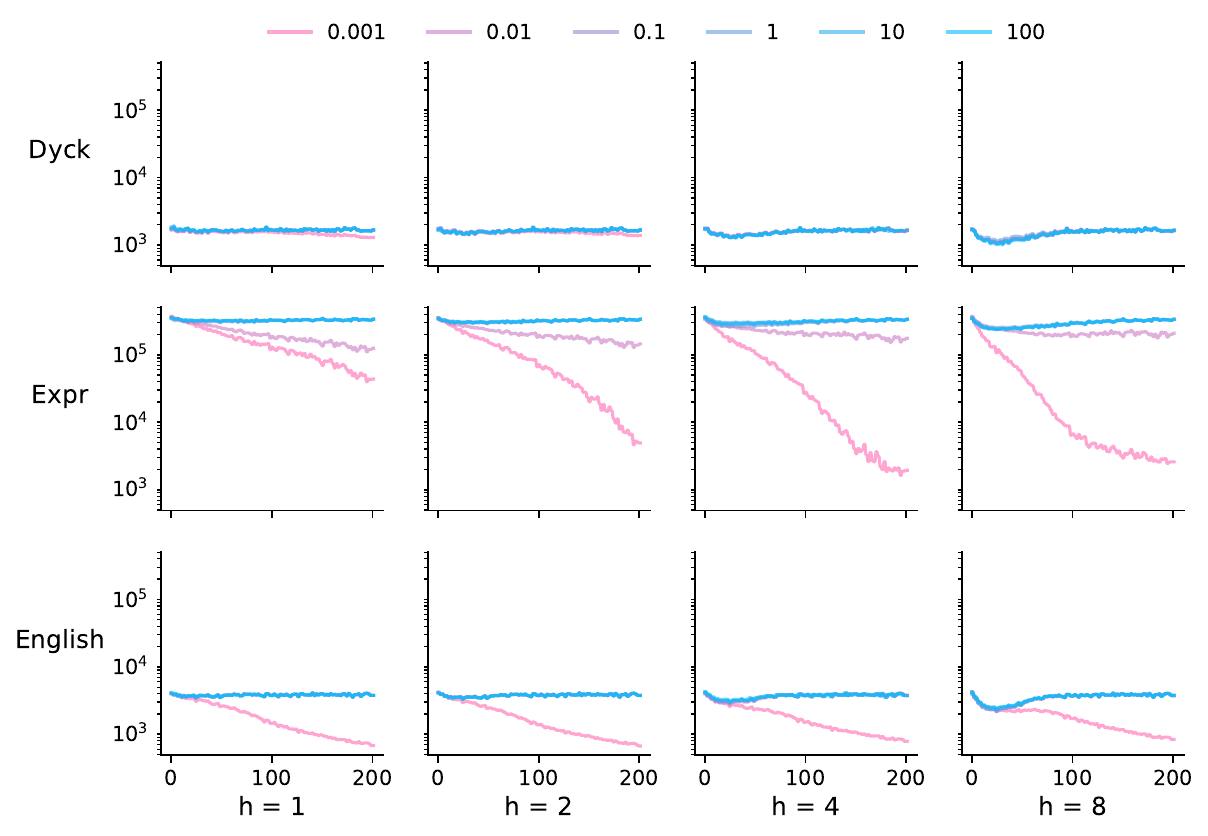}
    \caption{Reconstruction Loss}
    \end{subfigure}
    ~
    \begin{subfigure}[t]{0.45\textwidth}
        \includegraphics[width=1.0\linewidth]{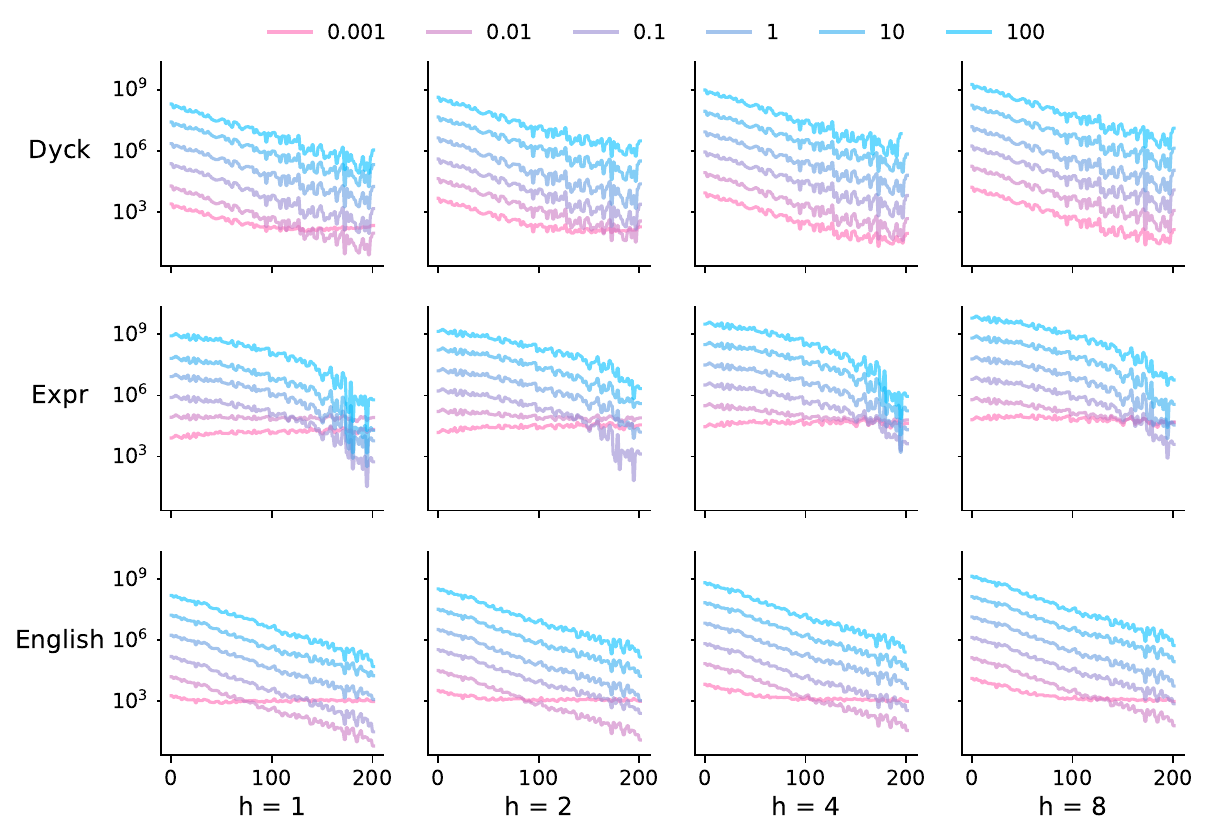}
        \caption{Regularization Loss}
    \end{subfigure}
    \caption{\textbf{$L_1$-regularized SAEs, without \texttt{pre\_bias} and without normalization.}}
    \label{fig:runset_1}
\end{figure*}

\begin{figure*}
    \centering
    \begin{subfigure}{0.45\textwidth}
    \includegraphics[width=1.0\linewidth]{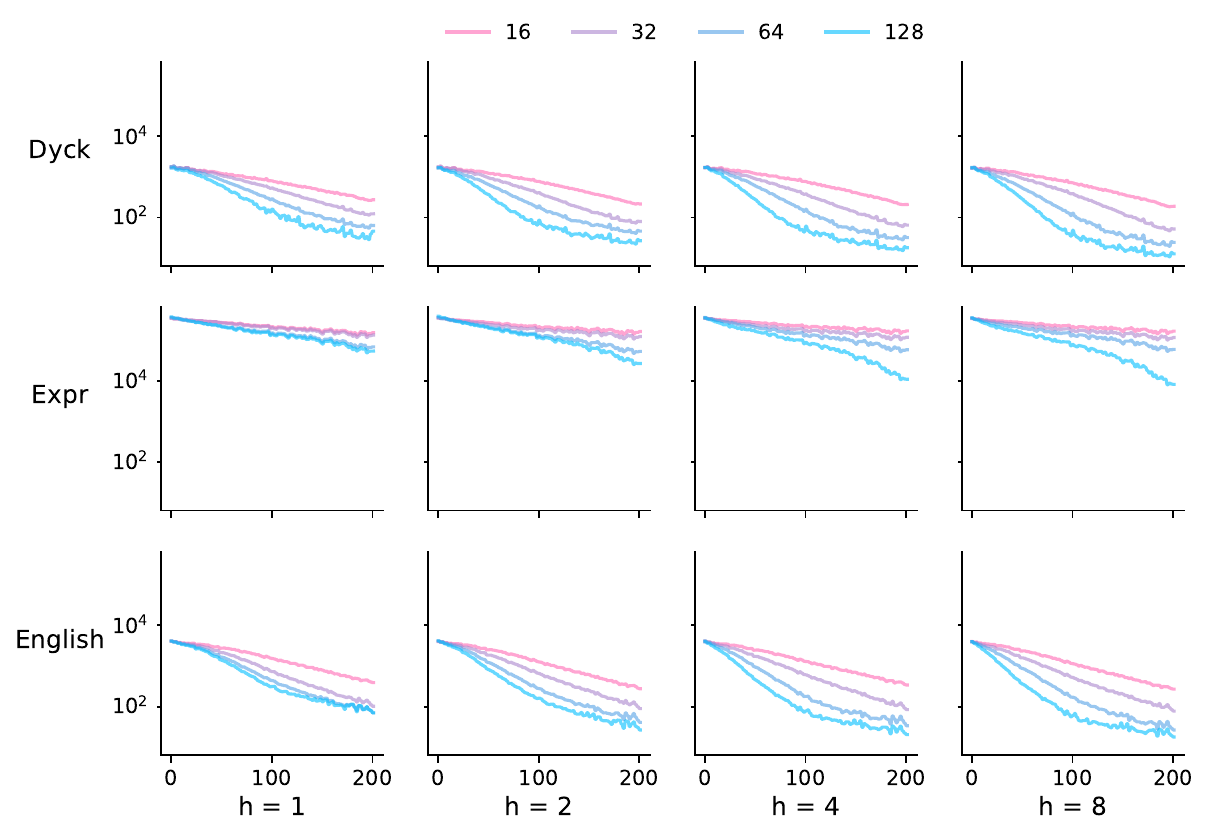}
    \caption{Train Loss (only reconstruction loss)}
    \end{subfigure}
    ~
    \caption{\textbf{top-$k$-regularized SAEs, without \texttt{pre\_bias} and without normalization.}}
    \label{fig:runset_2}
\end{figure*}

\begin{figure*}
    \centering
    \begin{subfigure}{0.45\textwidth}
    \includegraphics[width=1.0\linewidth]{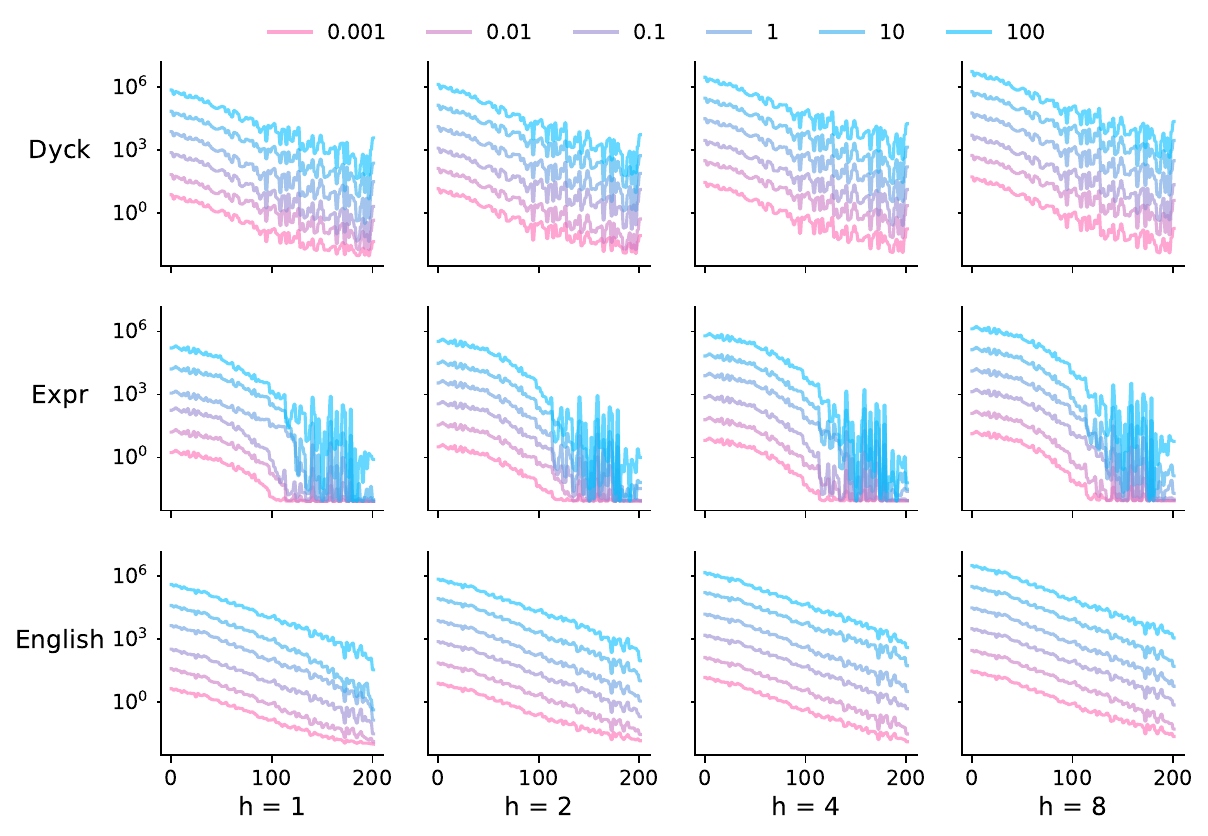}
    \caption{Train Loss (sum of reconstruction and regularization losses)}
    \end{subfigure}
    ~
    \vspace{5mm}
    \begin{subfigure}{0.45\textwidth}
    \includegraphics[width=1.0\linewidth]{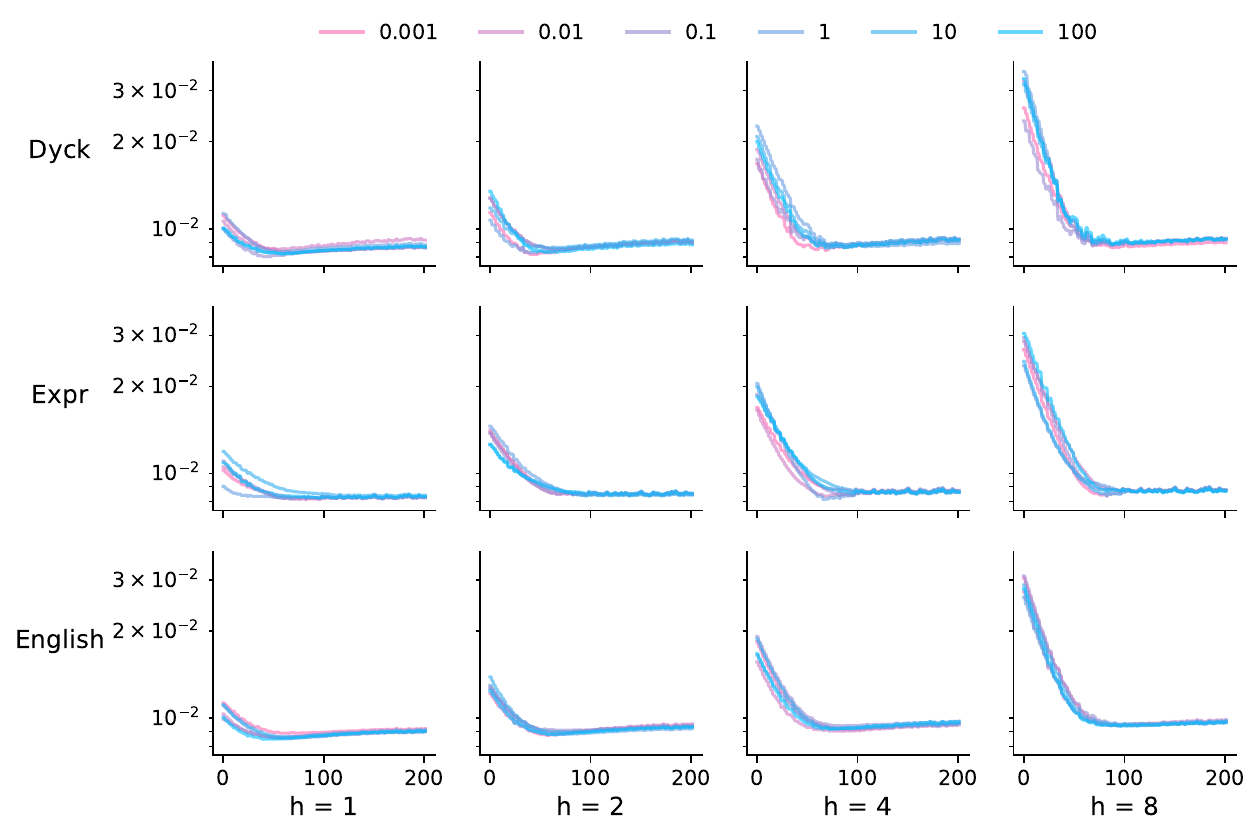}
    \caption{Reconstruction Loss}
    \end{subfigure}
    ~
    \begin{subfigure}{0.45\textwidth}
        \includegraphics[width=1.0\linewidth]{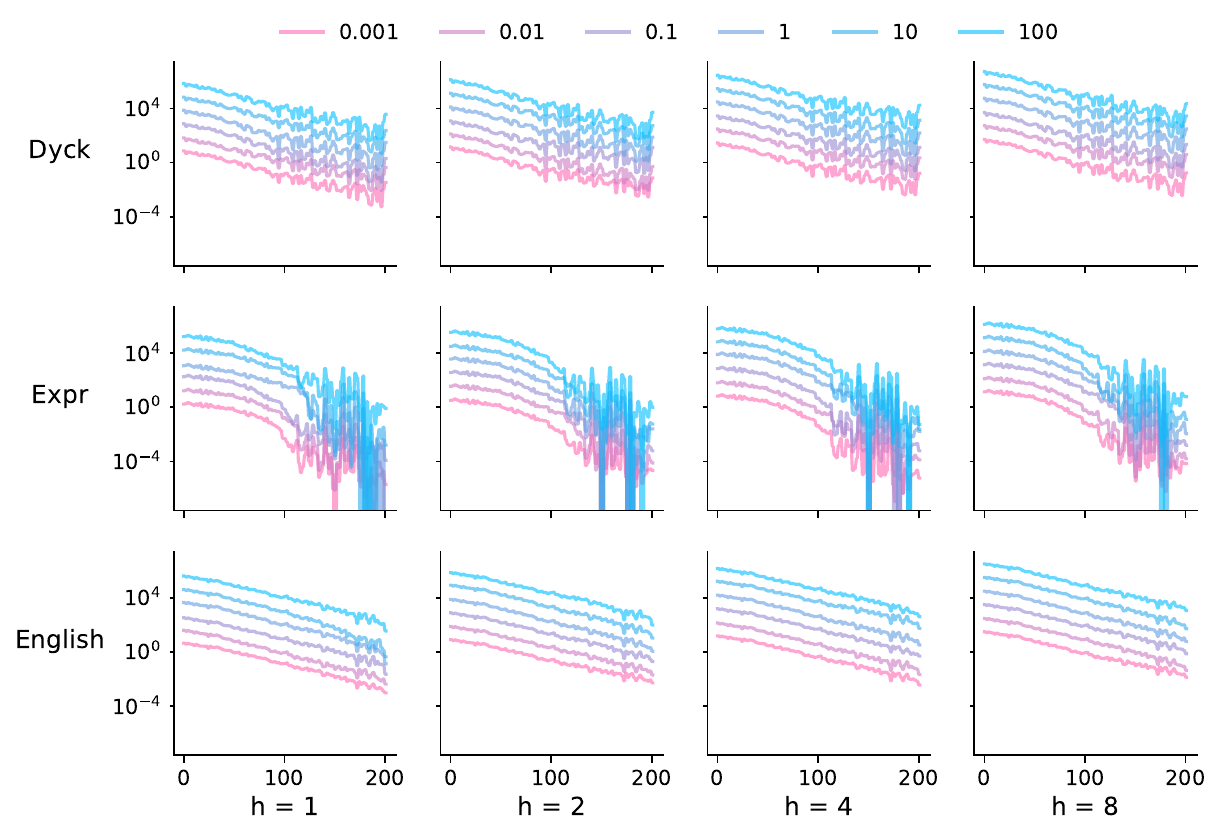}
        \caption{Regularization Loss}
    \end{subfigure}
    \caption{\textbf{$L_1$-regularized SAEs, with \texttt{pre\_bias} and normalized inputs and decoder directions.}}
    \label{fig:runset_3}
\end{figure*}

\begin{figure*}
    \centering
    \begin{subfigure}{0.45\textwidth}
    \includegraphics[width=1.0\linewidth]{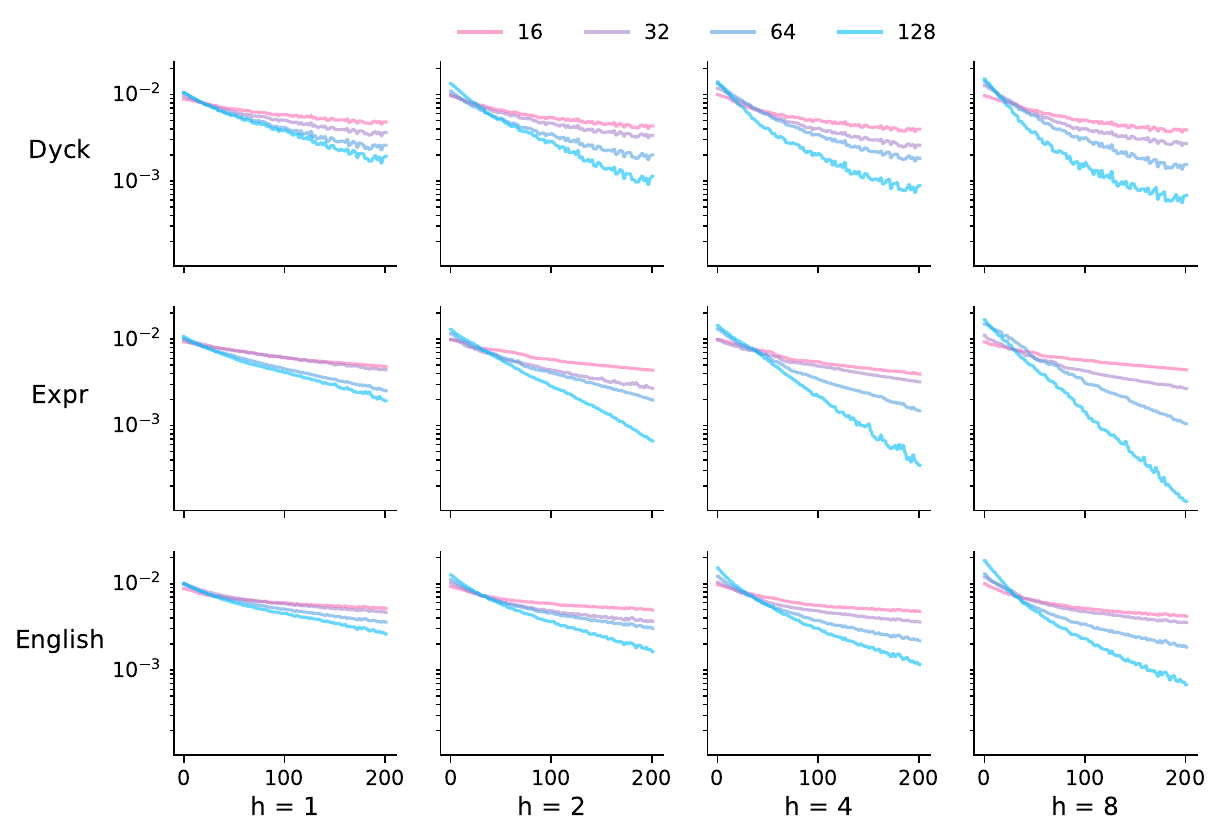}
    \caption{Train Loss (only reconstruction loss)}
    \end{subfigure}
    ~
    \caption{\textbf{top-$k$-regularized SAEs, with \texttt{pre\_bias} and normalized inputs and decoder directions.}}
    \label{fig:runset_4}
\end{figure*}

\begin{figure*}
    \centering
    \begin{subfigure}{0.45\textwidth}
    \includegraphics[width=1.0\linewidth]{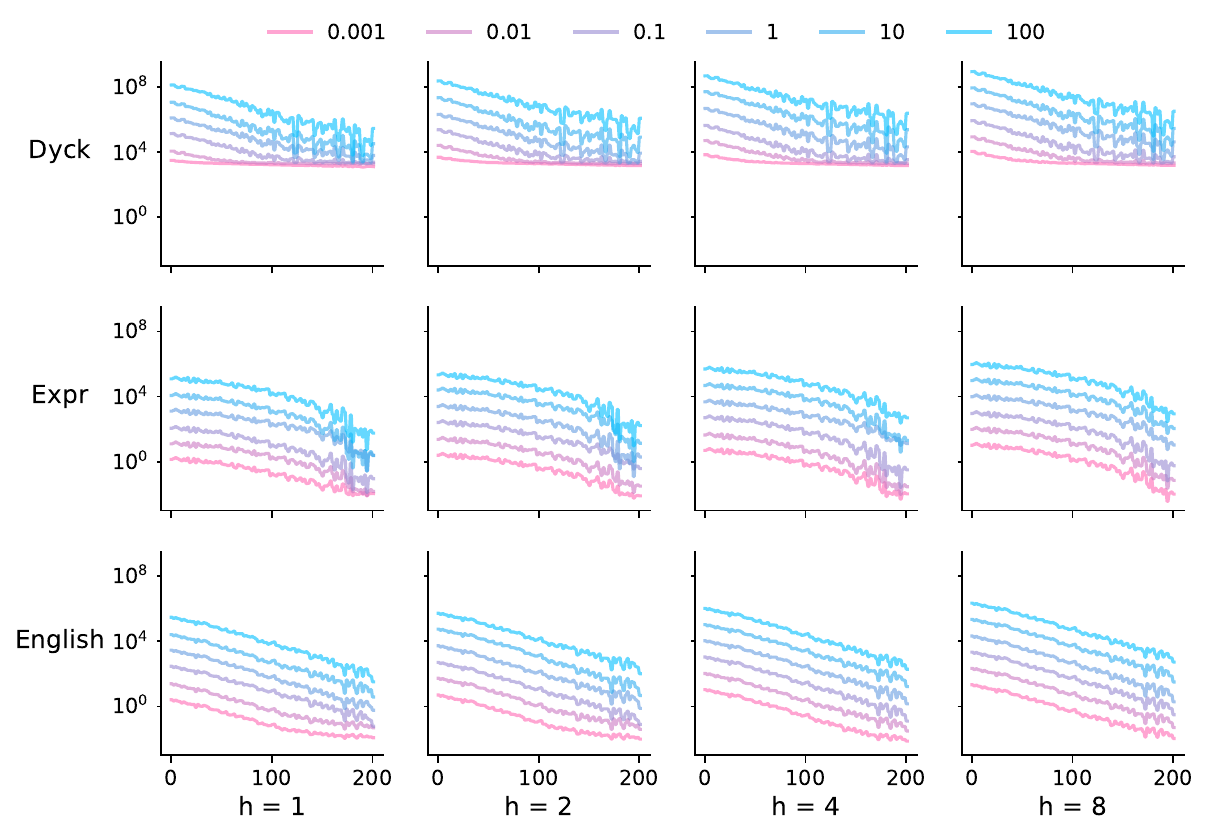}
    \caption{Train Loss (sum of reconstruction and regularization losses)}
    \end{subfigure}
    ~
    \vspace{5mm}
    \begin{subfigure}{0.45\textwidth}
    \includegraphics[width=1.0\linewidth]{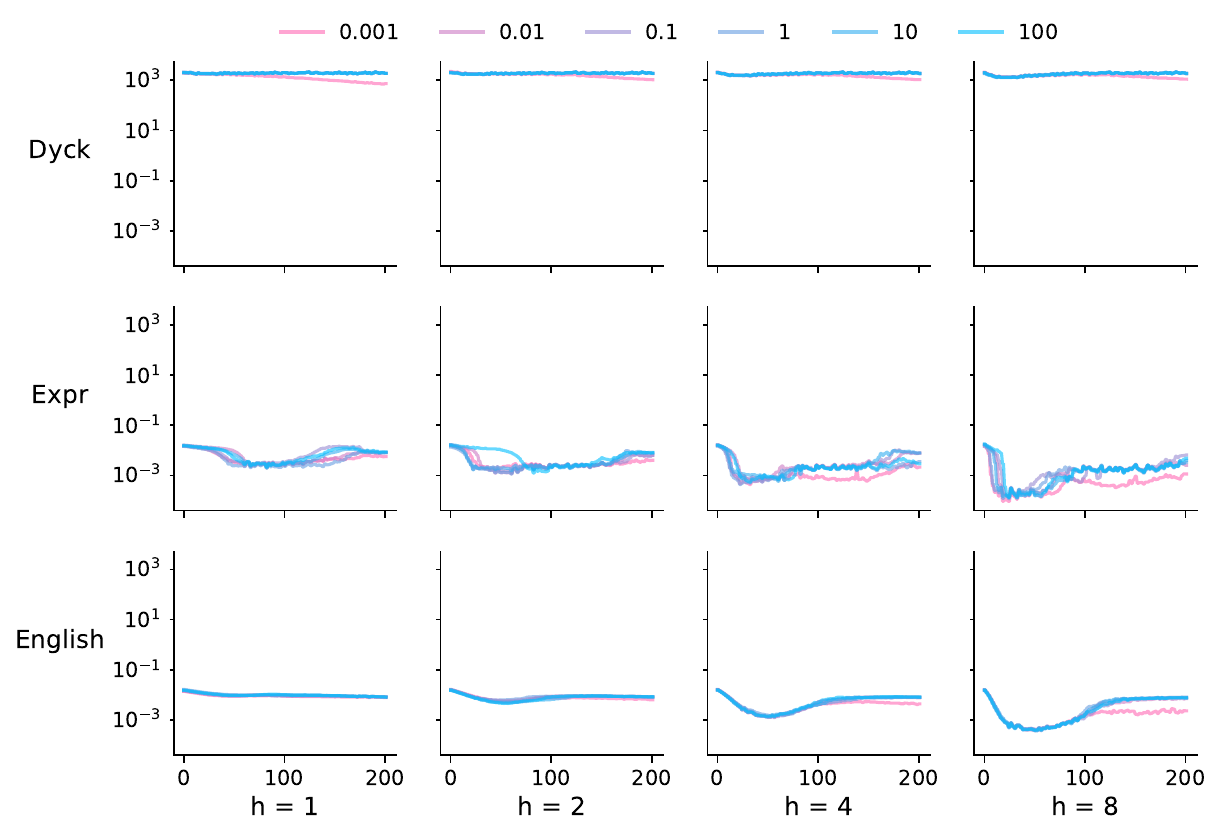}
    \caption{Reconstruction Loss}
    \end{subfigure}
    ~
    \begin{subfigure}{0.45\textwidth}
        \includegraphics[width=1.0\linewidth]{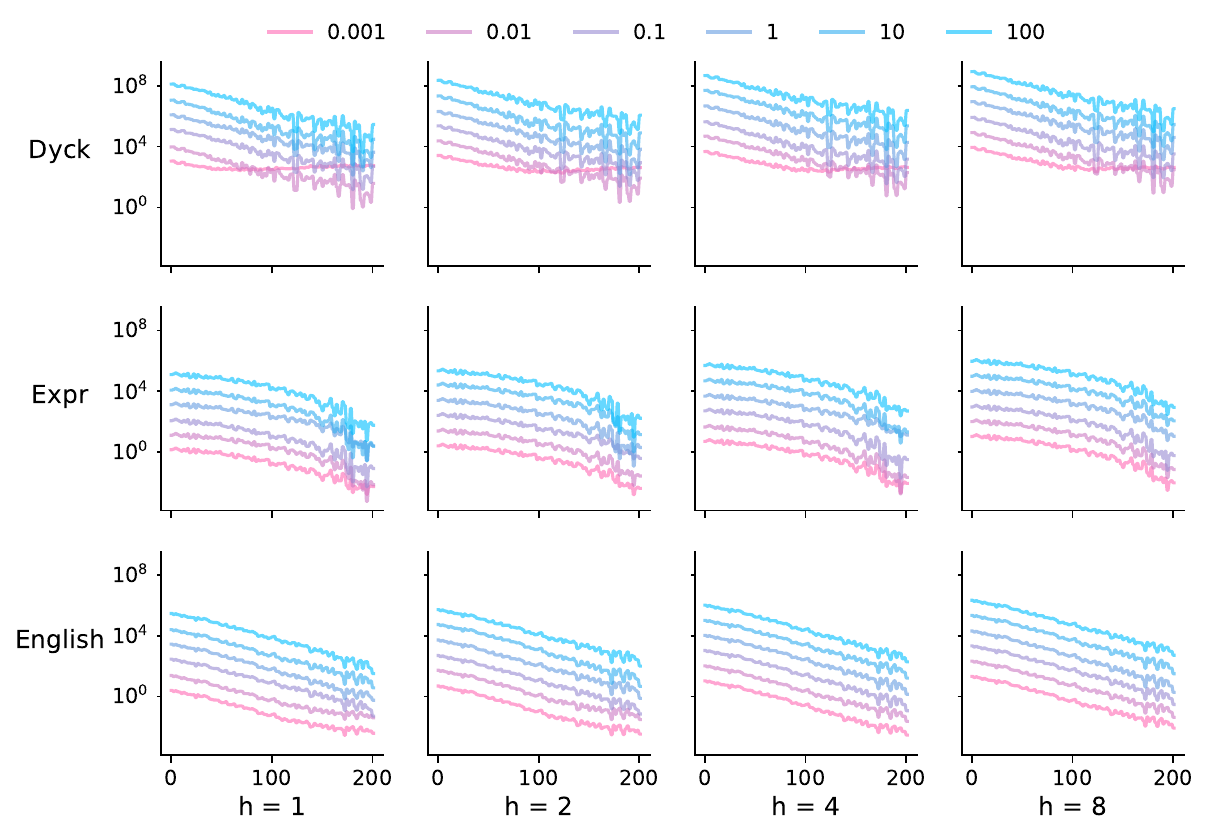}
        \caption{Regularization Loss}
    \end{subfigure}
    \caption{\textbf{$L_1$-regularized SAEs, without \texttt{pre\_bias} and normalized inputs.}}
    \label{fig:runset_6}
\end{figure*}

\begin{figure*}
    \centering
    \begin{subfigure}{0.45\textwidth}
    \includegraphics[width=1.0\linewidth]{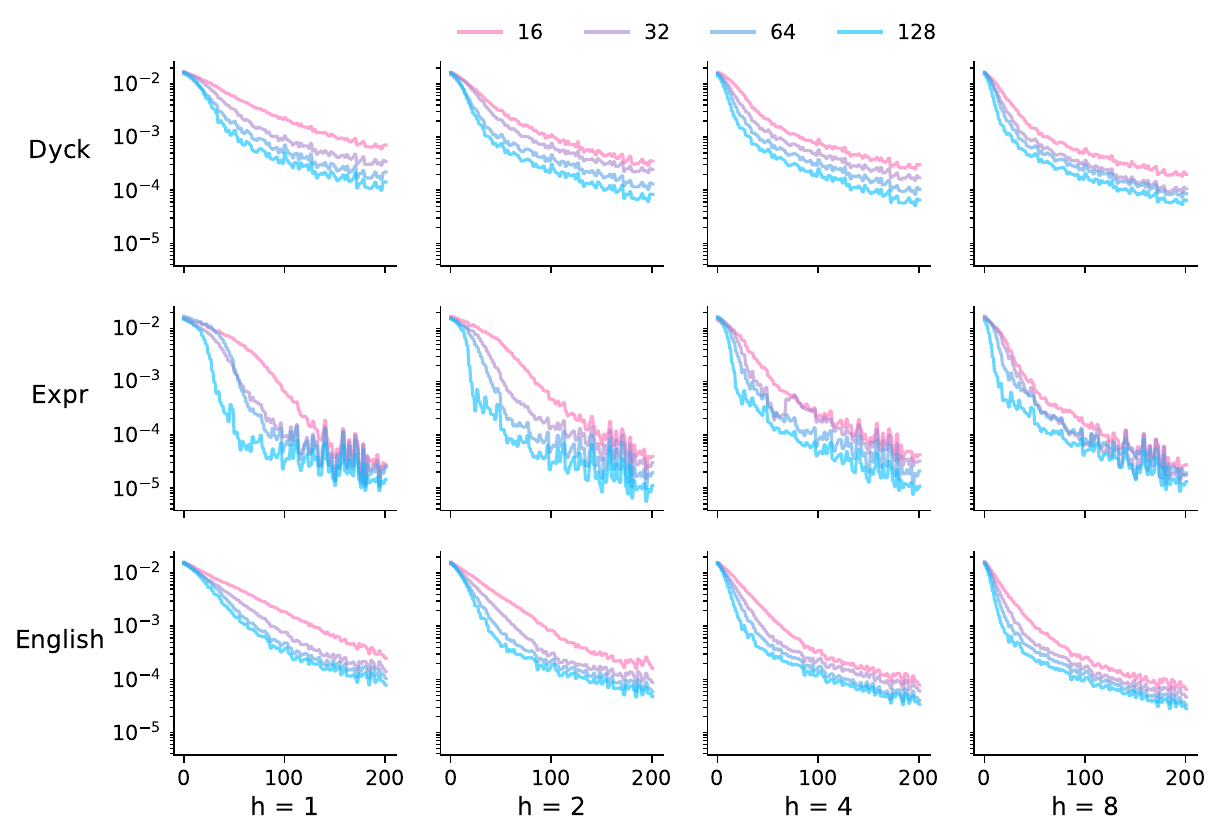}
    \caption{Train Loss (only reconstruction loss)}
    \end{subfigure}
    ~
    \caption{top-$k$-regularized SAEs, without \texttt{\textbf{pre\_bias} and normalized inputs.}}
    \label{fig:runset_8}
\end{figure*}

\begin{figure*}
    \centering
    \begin{subfigure}{0.45\textwidth}
    \includegraphics[width=1.0\linewidth]{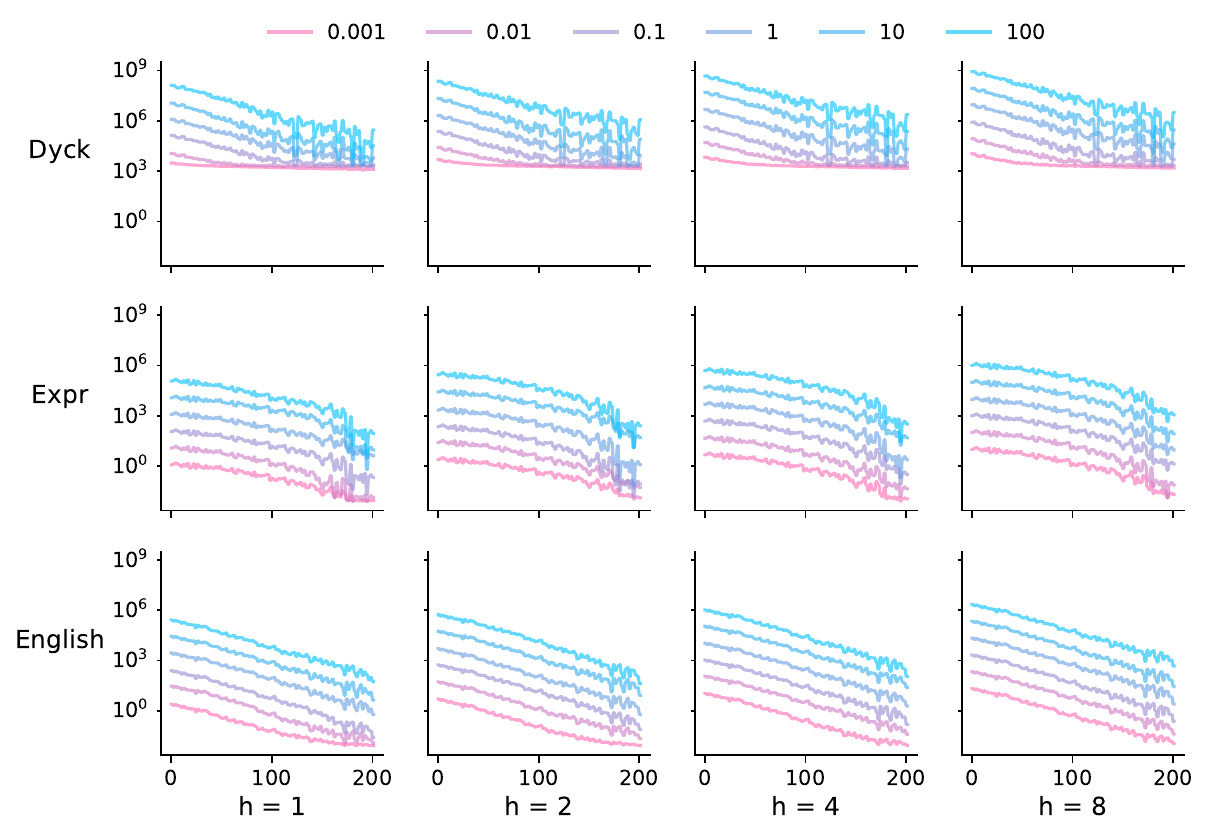}
    \caption{Train Loss (sum of reconstruction and regularization losses)}
    \end{subfigure}
    ~
    \vspace{5mm}
    \begin{subfigure}{0.45\textwidth}
    \includegraphics[width=1.0\linewidth]{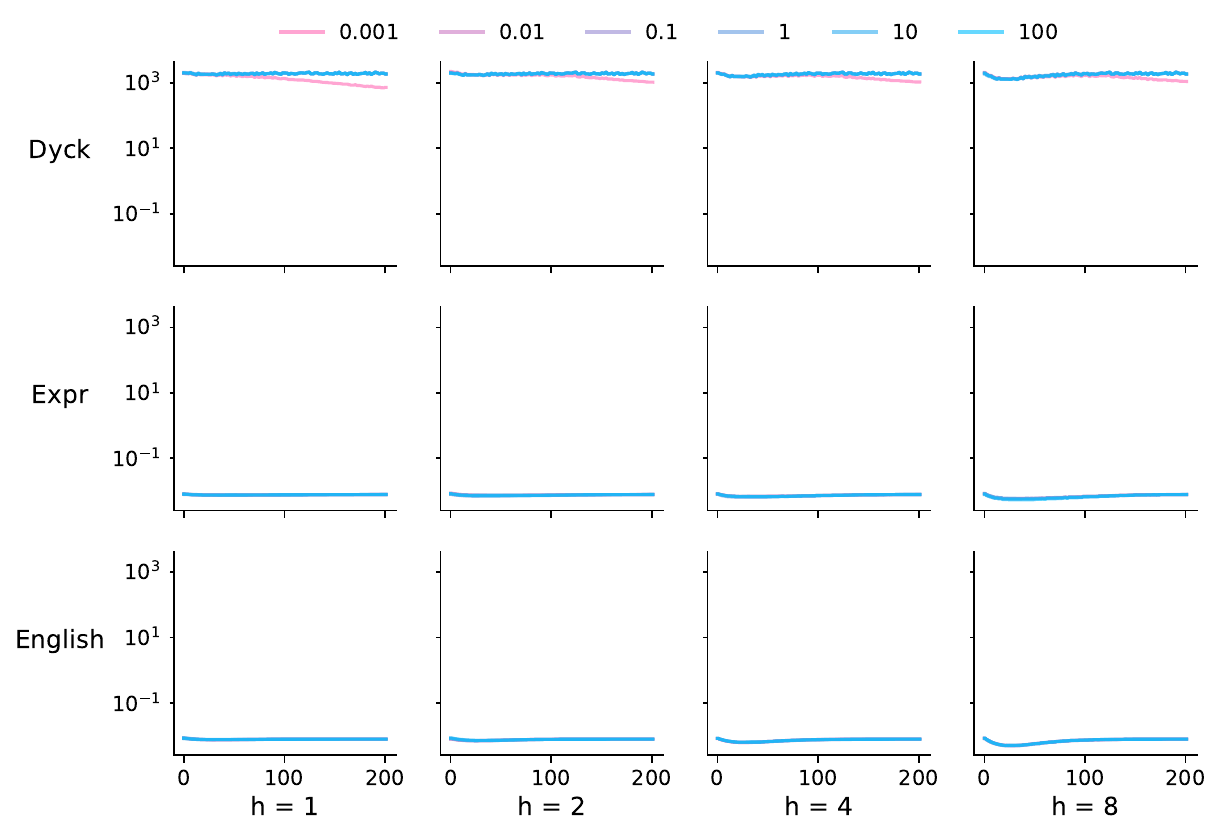}
    \caption{Reconstruction Loss}
    \end{subfigure}
    ~
    \begin{subfigure}{0.45\textwidth}
        \includegraphics[width=1.0\linewidth]{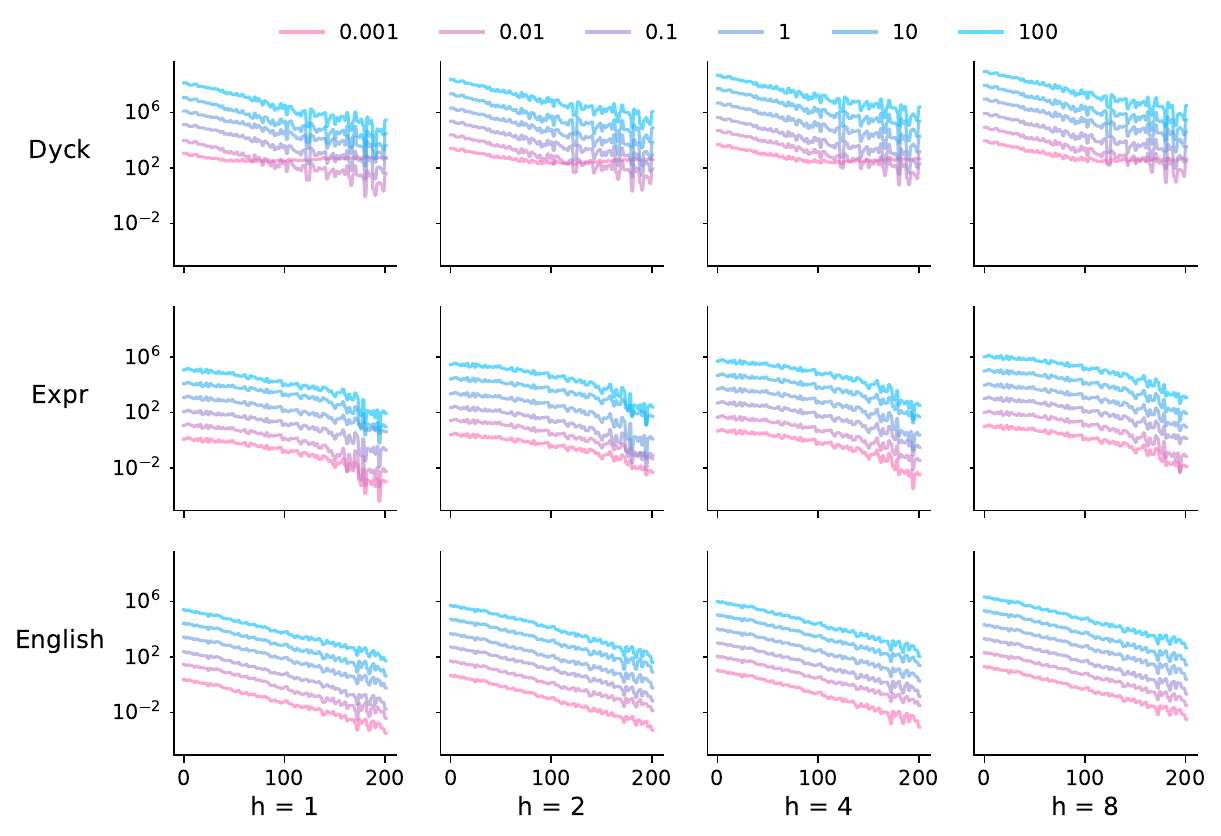}
        \caption{Regularization Loss}
    \end{subfigure}
    \caption{\textbf{$L_1$-regularized SAEs, without \texttt{pre\_bias} and normalized inputs and reconstructions.}}
    \label{fig:runset_7}
\end{figure*}

\begin{figure*}
    \centering
    \begin{subfigure}{0.45\textwidth}
    \includegraphics[width=1.0\linewidth]{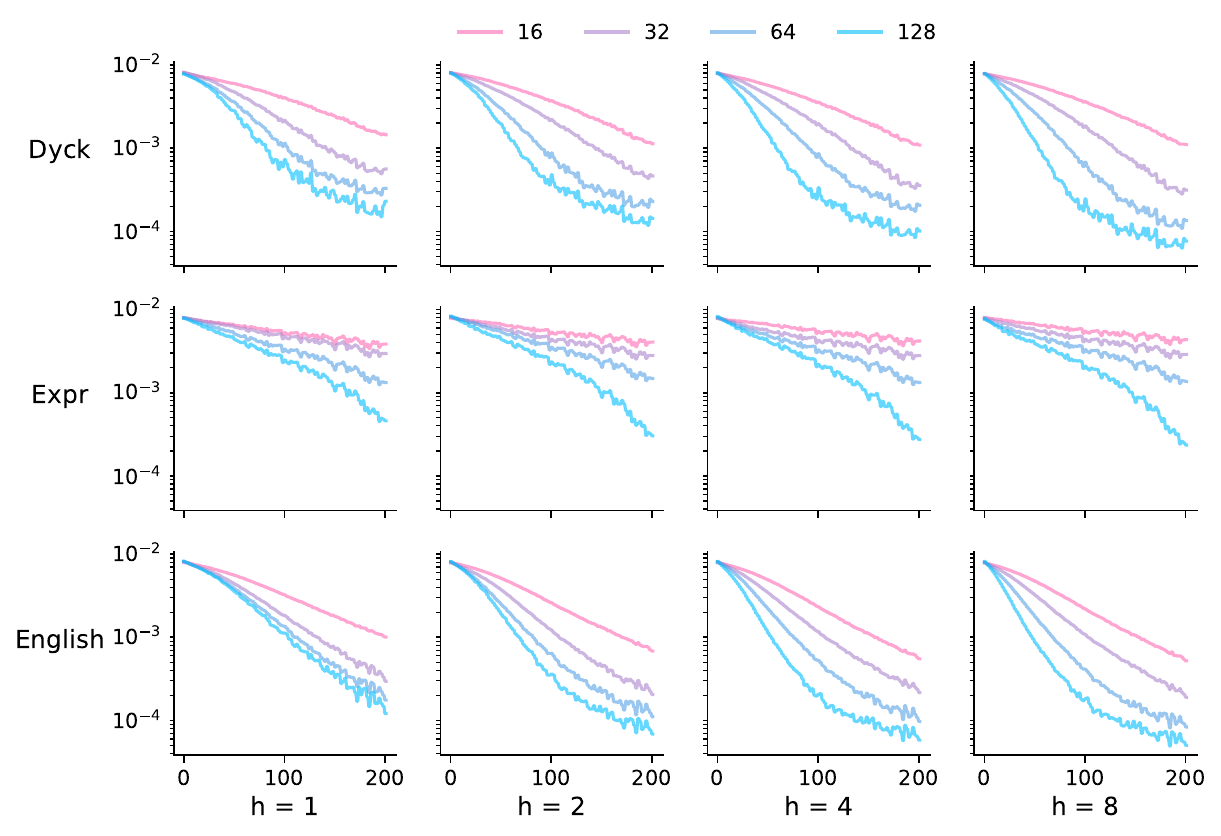}
    \caption{Train Loss (only reconstruction loss)}
    \end{subfigure}
    ~
    \caption{top-$k$-regularized SAEs, without \texttt{\textbf{pre\_bias} and normalized inputs and reconstructions.}}
    \label{fig:runset_9}
\end{figure*}

Figs.~\ref{fig:runset_1} to~\ref{fig:runset_9} show the details of the various hyperparameter settings we have considered for SAEs, along with plots of the losses during training. For convenience, we record these details in Tab.~\ref{table:runsets} as well.
Note that, in the case of $L_1$-regularized SAEs, the training loss is the sum of the reconstruction loss (the MSE loss between the SAE reconstructions and the inputs) and the regularization loss (the norm of the hidden representation, multiplied by the $L_1$ coefficient).
The SAEs are trained for 5000 iterations, with losses logged every 25 iterations.

The Transformer models are trained for $7 \times 10^4$ iterations.

\end{document}